\documentclass{article} 
\usepackage{iclr2027_conference,times}

\usepackage{amsmath,amsfonts,bm}

\def\eqref#1{equation~\ref{#1}}

\def\1{\bm{1}}

\DeclareMathAlphabet{\mathsfit}{\encodingdefault}{\sfdefault}{m}{sl}
\SetMathAlphabet{\mathsfit}{bold}{\encodingdefault}{\sfdefault}{bx}{n}

\usepackage{hyperref}
\usepackage{url}
\usepackage{multirow}
\usepackage[normalem]{ulem}
\useunder{\uline}{\ul}{}
\usepackage{booktabs}
\usepackage{graphicx}
\usepackage{amssymb}

\newcommand{\Name}{IMUDINO}

\title{\Name{}: Self-distilled Foundation Model for Inertial Sensing}

\author{%
    Zhoutong Ye$^{1\ddagger}$ \And
    Chengwen Zhang$^{1*}$ \And
    Zixuan Wang$^{1*}$ \And
    Xufeng Jian$^{1*}$ \AND
    Zhe He$^1$ \And 
    Qingyang Wan$^1$ \And
    Xutong Wang$^1$ \And
    Jiaqi Liu$^1$ \AND
    Mingze Sun$^1$ \And
    Yu Mei$^1$ \And
    Chun Yu$^{1\dagger}$ \And
    Yuanchun Shi$^1$ \AND
    \fontsize{9pt}{10.8pt}\selectfont
    ~~~~~~~~~~~~~~$^1$Tsinghua University $^\dagger$Corresponding Author $^\ddagger$Project Leader $^*$Equal Contribution \AND
    \fontsize{8pt}{10.8pt}\selectfont
    ~~~~~~~~~~~~~~~~~~~~~~~~~~~~~~~~~~~~~~~~~~~~~~~~~~~~~~~~~~~~~\texttt{yezt24@mails.tsinghua.edu.cn}
}

\iclrfinalcopy 
\begin{document}

\maketitle

\begin{abstract}
We present \Name{}, the first foundation model (FM) for inertial sensing to achieve competitive performance across diverse sensing tasks including human activity recognition (HAR), health, motion capture, and odometry. Prior inertial FMs often focus on one subfield of inertial sensing and sometimes underperform state-of-the-art supervised methods (e.g., in HAR), making their scope considerably narrower than FMs for vision and natural language. Moreover, they often require fixed input length and sampling rate, and generalize poorly to unseen sensor locations. To build a general-purpose foundation model for inertial sensing, we adapt the DINO self-supervised learning (SSL) paradigm to inertial data through a set of semantics-preserving motion transformations that simulate variations in motion pattern and sensor setup. For model architecture, we use a plain transformer encoder with frequency-aware positional encoding that natively supports varying input length and sampling rate. This simple setup proved remarkably effective for learning global and dense features generalizable across users, sensor setups, tasks, and, in the case of robot odometry, embodiments. We hope \Name{} can serve as a backbone for inertial sensing and kickstart the development of general-purpose inertial FMs. Code and checkpoints are available \href{https://github.com/Cambrian-yzt/IMUDINO-Release}{here}.
\end{abstract}

\section{Introduction}


Self-supervised foundational models (FMs) are core building blocks of modern natural language processing (NLP) and computer vision (CV). By removing the need for labeled data and not being tailored to a specific task, these models can leverage large-scale, easy-to-acquire unlabeled data and learn general-purpose features that can be readily used in downstream tasks. Similar to NLP and CV, inertial sensing covers diverse downstream tasks ranging from HAR to robot odometry, and would benefit immensely from a general-purpose foundation model that learns generalizable motion features from unlabeled IMU data. However, to truly fulfill this vision, an inertial FM must (1) perform competitively among task-specific supervised models on a variety of downstream tasks to justify their adoption, and (2) be versatile enough to support the wide range of input lengths, sampling rates, and sensor locations encountered in real-world scenarios. Existing work falls short in these regards, and a \textit{general-purpose foundation model} for inertial sensing remains an open problem.

To this end, we introduce \Name{}, the first FM with competitive performance on diverse inertial sensing tasks spanning HAR, health and gait analysis, motion capture, and robot odometry. \Name{} also provides native support for variable input length and sampling rate, and generalize to diverse sensor placements, which are often overlooked by existing inertial FMs. A key difference between \Name{} and previous work is the self-supervised learning (SSL) technique: instead of contrastive learning, masked reconstruction, or augmentation prediction, we adapt DINO self-distillation~\citep{caron2021dino} to inertial SSL. The core idea of DINO is encouraging different transformed views of the same sample to share a similar global feature representation. The transformation should therefore preserve the global semantics of the sample. For inertial sensing, we propose a set of \textit{semantics-preserving motion transformations}, including cropping, scaling along the magnitude and temporal axes, $SO(3)$ rotation, resampling, and input channel masking. This allows the model to learn a global representation that generalizes across different sensor orientation, user-specific motion patterns, and input channel availability (e.g., acc-only or full 6-axis IMU). We further adopt the iBOT++ patch-level training objective~\citep{cao2026tipsv2} to learn fine-grained dense features essential for tasks like motion capture and robot odometry. 

The effectiveness of our training objective allows \Name{} to outperform previous methods while using a simple model architecture and orders of magnitudes less training data ($\sim$500 vs $\sim$700000 days used by~\cite{yuan2024self, xu2025relcon}). Specifically, we employ a minimalist transformer architecture augmented only by \textit{frequency-aware positional encoding}, which scales the frequency base with the input's sampling rate and enable the model to learn relational patterns rooted in physical time instead of token index. For training, we use a filtered subset curated from three public datasets, namely Capture24~\citep{chan2024capture24}, SHL~\citep{gjoreski2017shl}, and PAAWS~\citep{potter2025paaws}. These datasets are collected from various positions (phone, wrist, thigh, ankle, waist, etc.), contributing to \Name{}'s generalization capabilities with regard to sensor location.

We evaluate \Name{} on a diverse set of downstream tasks using its frozen features. In HAR, \Name{} outperformed supervised HAR models, HAR FMs, and general-purpose time series FM baselines on 7 datasets covering diverse activities and heterogeneous sensor setups. In health and gait analysis, \Name{} features can be readily used for activity intensity estimation, fall detection, and gait recognition. In inertial motion capture, \Name{} achieves competitive performance on 4 different human pose estimation datasets with a simple MLP probe and no physical priors. Finally, in robot odometry, \Name{} outperforms Tartan IMU, a SOTA robot odometry model. Together, these results prove that \Name{} can extract both global and dense features that can generalize across users, sensor setups, tasks, and, in the case of robot odometry, embodiments, making \Name{} the first general-purpose FM for the broader inertial sensing community. 

\section{Related Work}

\subsection{Foundation Models for Inertial Sensing}

The quest for a foundation model for inertial sensing started with self-supervised HAR. Early examples include \citet{saeed2019multitask}, an augmentation prediction method, and LIMU-BERT~\citep{xu2021limubert}, which mirrors the masked reconstruction objective of BERT. In addition, \citet{tang2020exploring} brought contrastive learning into the field by adapting SimCLR~\citep{chen2020simclr} to IMU HAR. Work on inertial FMs have since followed the paths set by these three early attempts. For instance,~\citet{yuan2024self} expanded augmentation prediction to large scale pretraining data. REBAR~\citep{xu2023rebar} introduced a two-stage solution, training the model first to reconstruct retrieved motifs and then with contrastive loss. RelCon~\citep{xu2025relcon} brought motion similarity into contrastive learning. HAR-FM~\citep{qiu2025harfm} combines the contrastive and masked prediction objectives. However, the results here are mixed: progress in this line of work is mainly measured by improved HAR metrics, yet SSL methods like RelCon struggle to outperform early fully-supervised models like DeepConvLSTM~\citep{deepconvlstm} and simple ConvNets, let alone more modern supervised HAR methods like TSF~\citep{zhang2026tsf}. Apart from the narrow scope and uncompetitive performance, many of these models (e.g., LIMU-BERT, \citet{yuan2024self}, RelCon, etc.) requires fixed input length and sampling rate, limiting their potential in diverse scenarios. Entirely unrelated to this family of work, there is also TartanIMU~\citep{zhao2025tartanimu}, an FM for inertial odometry in robots. A key difference between the HAR FMs and TartanIMU is that robot odometry relies on \textit{dense features}, while HAR and health tasks use \textit{global features} only.

In summary, the field of inertial FMs remains fragmented as most models only focus on one subfield of inertial sensing. There is also a disconnect between FMs for global features and dense features. The result is a myriad of narrowly-scoped models akin to a vision ``FM'' useful only for image classification. We argue that a more general-purpose inertial FM supporting global and dense prediction is both desirable and possible. This is the motivation behind \Name{}, the first inertial FM with broad support for both global tasks (HAR, health) and dense prediction (MoCap, robot odometry).

\subsection{Why DINO for Inertial Sensing?}

DINO~\citep{caron2021dino} is a highly successful SSL paradigm in CV. The core objective of DINO is aligning the global representation (i.e., the class token) of transformed views of the same image. \citet{zhou2021ibot} proposed the iBOT patch-level masked token reconstruction objective to complement DINO's global objective. On top of iBOT, DINOv2~\citep{oquab2024dinov2} added the KoLeo regularization loss to encourage uniform spread of the learned features, and massively scaled up the training data. DINOv3~\citep{simeoni2025dinov3} introduced the Gram anchoring technique for improving dense feature quality during very long training runs to enable further scaling. DINOv2 and DINOv3 have since been used as the backbone in various downstream tasks. Influential examples include DepthAnything~\citep{yang2024depthanything} for depth estimation and AnyLoc for localization~\citep{keetha2023anyloc}. DINOv3 further demonstrated SOTA capabilities when swapped into existing architectures for object detection, semantic segmentation, and 3D understanding~\citep{simeoni2025dinov3}. Finally, the DINOv2 and DINOv3 SSL recipes have also been successfully adapted to satellite imagery~\citep{tolan2024very, simeoni2025dinov3} and medical imaging~\citep{vorontsov2024foundation}.

DINOv2's training objective combines the strengths of contrastive methods and masked autoencoder (MAE)~\citep{he2022mae}, which are common approaches to inertial SSL. It provides both global representations with semantic view invariance and fine-grained dense features. Moreover, DINOv2's anti-collapse mechanisms (e.g. EMA teacher, centering and sharpening, KoLeo, etc.) apply to model weights and features, not the input, making them modality-agnostic and easily transferable to inertial SSL. In this work, we adapt the DINOv2 training objective to inertial sensing by generating views through semantics-preserving motion transformations. We demonstrate through experiments that \Name{}, like its CV counterparts, can be integrated into a wide variety of inertial sensing tasks and achieved competitive performance, significantly broadening the scope of inertial FMs.
\section{\Name{}}
\label{sec:method}

\subsection{Training Objective}

\begin{figure*}[htbp]
  \centering
  \includegraphics[width=0.8\textwidth]{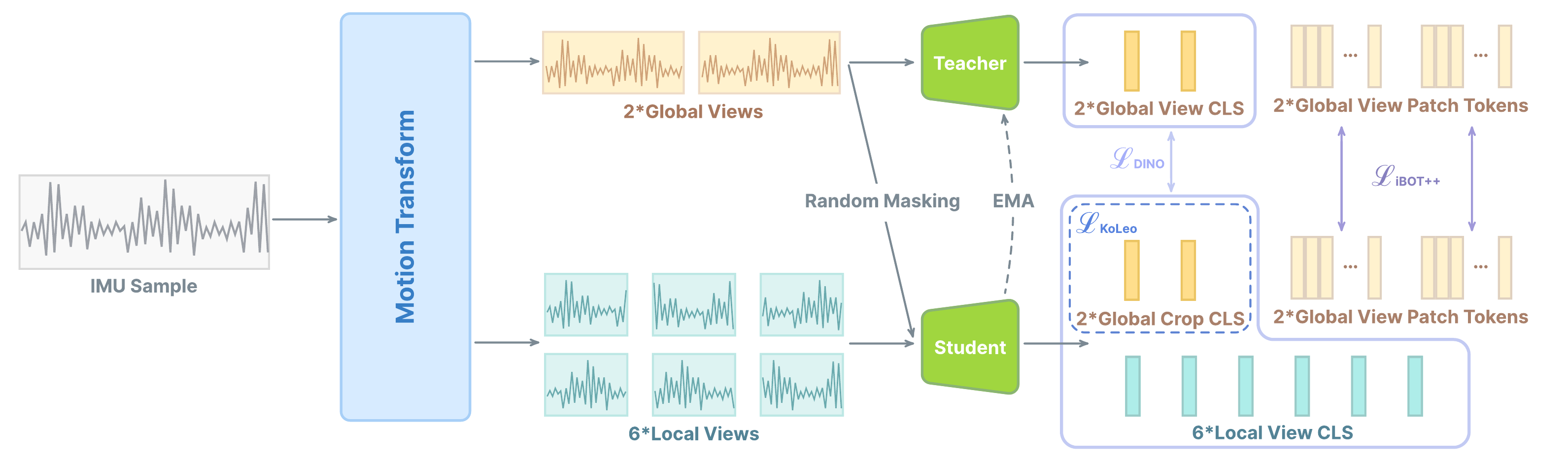}
  \caption{An overview of the \Name{} training objective.}
  \label{fig:method}
\end{figure*}

The training objective of \Name{} has three components: the global objective ($\mathcal{L}_{\mathrm{DINO}}$), the patch-level objective ($\mathcal{L}_{\mathrm{iBOT++}}$), and a KoLeo regularization loss ($\mathcal{L}_{\mathrm{KoLeo}}$). The student and its DINO and iBOT heads are optimized with backpropagation, whereas the teacher is the EMA of the student.

\textbf{Global objective.} This objective concerns the global features (i.e., the class token) extracted by a student and a teacher network, which are given transformed views of the same IMU sample. The student features are passed through a DINO head, which is an MLP that outputs \textit{prototype scores}. Softmax is then applied to the prototype score vector to obtain $p_s$. Similarly, we obtain the teacher's prototype scores using its own DINO head. We then apply sharpening and Sinkhorn-Knopp centering in DINOv2 fashion to obtain $p_t$. The global loss term is therefore:

$$
\mathcal{L}_{\mathrm{DINO}}=-\sum{p_t \log p_s}
$$

\textbf{Patch-level objective.} As in DINOv2, we apply the iBOT patch-level objective. We mask some of the input patches given to the student, while the same patches remain visible to the teacher. The masked student patch tokens are passed through the student iBOT head (identical in structure to the DINO head). The teacher's corresponding visible patch tokens are processed by the teacher iBOT head. As in $\mathcal{L}_{\mathrm{DINO}}$, we apply centering and softmax before calculating the patch-level loss term:

$$
\mathcal{L}_{\mathrm{iBOT}}=-\sum_{i}{p_{ti} \log p_{si}}
$$

where $i$ are patch indices of masked tokens. TIPSv2 further expanded this loss term to patch token pairs that are visible to \textit{both} the student and the teacher, leading to iBOT++. This anchors the student's patch representation to that of the teacher and encourages the preservation of local semantics. The only difference between $\mathcal{L}_{\mathrm{iBOT}}$ and $\mathcal{L}_{\mathrm{iBOT++}}$ is that $i$ loops over the indices of \textit{all} tokens in iBOT++, regardless of whether a token is masked for the student. 

\textbf{KoLeo regularizer.} Following DINOv2 and DINOv3, we use the KoLeo regularizer to encourage the features in a batch to spread uniformly in the feature space. The features used here are the $\ell_2$-normalized class tokens. Given $n$ feature vectors $(x_1, ...,x_n)$, the loss is defined as

$$
\mathcal{L}_{\mathrm{KoLeo}} = -\frac{1}{n}\sum_{i=1}^{n}{\log{(\min_{j \neq i}{||x_i - x_j||}})}
$$

Combining all three objectives, we arrive at the full \Name{} training objective:

$$
\mathcal{L} = \lambda_{d}\mathcal{L}_{\mathrm{DINO}} + \lambda_{i}\mathcal{L}_{\mathrm{iBOT++}} + \lambda_{k}\mathcal{L}_{\mathrm{KoLeo}}
$$

\subsection{Semantics-preserving Motion Transformation}
\label{sec:method-transforms}

The transformations applied to the views reflect what the \textit{global representation} should be robust against. Therefore, we choose the following set of \textit{semantics-preserving motion transformations} from the perspective of desirable global invariance. See Appendix~\ref{sec:appendix-impl-details-transform} for detailed implementation.

\textbf{Cropping.} Similar to the vision modality, we apply random cropping to generate views. For IMU data, this is grounded in the temporal continuity of human motion: actions happening at around the same time are usually related and semantically similar, and should therefore share similar feature representations. Following DINO, we include 2 types of cropped views: \textit{global views} that cover $0.5\times$ to $0.8\times$ the length of a sample, and \textit{local views} that cover $0.05\times$ to $0.4\times$ the length of a sample. The idea here is to encourage local-global correspondence in the learned representation.

\textbf{Scaling.} Scaling is a commonly used data augmentation technique for inertial data~\citep{ye2026activityseeker}. In \Name{}, we randomly apply scaling along two axes: the magnitude axes controlling the motion's intensity, and the temporal axes controlling how fast the forces and movements change. This is grounded in the fact that semantically similar motions are performed at different intensity and pace by different people or under different contexts.

\textbf{Rotation.} We apply random $SO(3)$ rotations to the views to simulate changes in sensor orientation. This encourages the learned global representation to be less about which direction the sensor is facing and more about the underlying motion itself.

\textbf{Resampling.} Whether a motion is recorded at 50Hz or 200Hz does not change its global semantics. Therefore, we apply random resampling to make \Name{} be robust and consistent when faced with the wide range of sampling rates encountered in real-world applications.

\textbf{Channel masking.} We apply random channel masking so that the learned representation is not over-reliant on a certain input channel. This is also useful for adapting \Name{} to downstream tasks where only triaxial acceleration input is available.

\subsection{IMUFormer: A Minimalist Transformer for Inertial Data}
\label{sec:method-imuformer}

The transformer architecture has seen increasing adoption in inertial sensing. Beginning with a vanilla transformer encoder~\citep{shavit2021boosting}, IMU transformer architectures have since become increasingly complex~\citep{zhang2022ifconvtransformer, zhang2026tsf}. To demonstrate that the strong performance of \Name{} is driven by our effective SSL setup and not complicated model architecture, we return to the basics in this work. Starting from the standard ViT model, we make only two adaptations. First, we replace the 2D image ``patchify'' layer with an 1D non-overlapping convolution that project the original IMU frames into patch tokens. Second, we replace the learned absolute positional encoding (PE) with a frequency-aware sinusoidal PE that scales sequence positions according to each sample’s sampling frequency. For a sample with frequency \(f_b>0\), position \(p\) is mapped to \(\tilde p_b=p f_{\mathrm{ref}}/f_b\), where \(f_{\mathrm{ref}}\) is a fixed reference frequency. The encoding is defined as

$$
\mathrm{PE}_{b,p,2i}=\sin\!\left(\tilde p_b\,\beta^{-2i/D}\right),
\qquad
\mathrm{PE}_{b,p,2i+1}=\cos\!\left(\tilde p_b\,\beta^{-2i/D}\right),
$$

We empirically set \(\beta=100\) and \(f_{\mathrm{ref}}=100\mathrm{Hz}\). This parameter-free formulation assigns identical positional encodings to equal elapsed times across sampling frequencies. This, in conjunction with the resampling transformation during training, enables \Name{} to generalize to any reasonable IMU sampling rate. This minimalist architecture is referred to as \textit{IMUFormer} in this paper.

\subsection{Training Data}
We train \Name{} on a dataset curated from three public wearable inertial datasets covering various sensor locations - Capture24~\citep{chan2024capture24}, SHL~\citep{gjoreski2017shl} and PAAWS-Lab~\citep{potter2025paaws}. The entire train set contains 11.6k hours of 6-axis IMU data. This is orders of magnitude smaller than the 700k - 800k person-days used in RelCon and~\citet{yuan2024self}.

\textbf{Capture24} consists of 3883 hours of wrist-worn accelerometer data collected from 151 free-living participants. We filtered out data with null, incomplete or inconsistent annotation, resulting in 2410 hours of triaxial acceleration data. We then zero-fill the non-existent gyroscope channels.

\textbf{SHL} is a smartphone IMU dataset covering 4 phone locations (hip, torso, hand, bag). The train and validation sets from the SHL2026 challenge, totaling 1138 hours of 6-axis IMU data, is used. 

\textbf{PAAWS-Lab} is collected from a controlled lab environment. The 252 participants contributed 4 hours of data each. 20 IMUs were placed on the participants' body. Specific locations include the waist, wrists, thighs and ankles. We filtered out static segments of the data (See Appendix~\ref{sec:appendix-pretrain-data-proc} for details) to improve training efficiency, resulting in 8069 hours of 6-axis IMU data.

\subsection{Implementation}

We divide the training set into samples using non-overlapping 1024-frame sliding windows. Each sample is then transformed into 2 global views and 6 local views per Sec.~\ref{sec:method-transforms}. The teacher receives the 2 global views as input, while the student receives all 8 views. The global objective $\mathcal{L}_{\mathrm{DINO}}$ is computed between the student views and teacher views in the exact same manner as DINOv2. For the patch-level objective $\mathcal{L}_{\mathrm{iBOT++}}$, we randomly select 50\% of student global views in a minibatch and replace a random ratio (sampled from $U(0.1,0.5)$) of patch tokens with a learned mask token. Note that $\mathcal{L}_{\mathrm{iBOT++}}$ applies to both masked and visible tokens. The $\mathcal{L}_{\mathrm{KoLeo}}$ regularization loss is computed over the 2 global view features generated by the student. We empirically set the loss weights to $\lambda_{d}=\lambda_{i}=1.0$ and $\lambda_{k}=0.2$. IMUFormer models of two sizes are used in the main experiments. A 6M parameter version comparable in size and FLOPs to most supervised and inertial FM baselines, and a larger 85M parameter version to demonstrate the scaling potential of our approach. Details about the models and training recipe are in Appendix~\ref{sec:appendix-impl-details-model-training}.

\section{Experiments}


We evaluate \Name{}'s performance as a general-purpose inertial FM in four steps. First, we evaluate HAR performance using the global features, a quintessential inertial FM task. Second, we explore using \Name{}'s dense features for inertial motion capture, an inherently challenging task. Third, we test \Name{}'s emergent cross-embodiment transfer capability with robot inertial odometry. Finally, we dissect how our training objective, data augmentation, and the IMUFormer model contribute to performance in an ablation study.

\subsection{Baselines}

For HAR, commonly used to evaluate inertial FMs, we compare \Name{} to SimCLR~\citep{chen2020simclr}, \citet{yuan2024self}'s method, RelCon~\citep{xu2025relcon}, and HAR-FM~\citep{qiu2025harfm}. To facilitate fair comparison and minimize the influence of training data and model architecture, we train all inertial FM baselines using \Name{}'s training set and the IMUFormer-6M backbone. For baselines that were originally implemented for acceleration-only data, we add 3 input channels for gyroscope. We further include two general-purpose time-series foundation models (TSFMs) as baselines: TimesFM~3.0~\citep{das2024timesfm} and Chronos-2~\citep{ansari2025chronos2} (330.7M \& 120M parameters, respectively), and use the official weights off-the-shelf. We also include task-specific supervised baselines in our comparison, namely DeepConvLSTM~\citep{deepconvlstm}, Attend \& Discriminate~\citep{abedin2021attenddiscriminate}, IF-ConvTransformer~\citep{zhang2022ifconvtransformer} and TSF~\citep{zhang2026tsf}, which represent the model-driven SOTA for IMU HAR. For inertial motion capture, DIP~\citep{huang2018dip}, TransPose~\citep{yi2021transpose}, TIP~\citep{jiang2022tip}, PIP~\citep{yi2022pip} and DynaIP~\citep{zhang2024dynamic} are used as the supervised baselines. For robot inertial odometry, we compare against Tartan IMU~\citep{zhao2025tartanimu}, the state-of-the-art model for that task.


\subsection{HAR: \Name{} has Exceptional Global Motion Descriptors}
\label{sec:exp-har}

We first evaluate \Name{} on seven HAR datasets: MotionSense~\citep{malekzadeh2018motionsense}, HAPT~\citep{reyes-ortiz2016hapt}, HHAR~\citep{stisen2015hhar}, DSADS~\citep{barshan2014dsads}, OPPORTUNITY~\citep{chavarriaga2013opportunity}, \citet{shoaib2014shoaib}, and WEAR~\citep{bock2024wear}. For all datasets except WEAR, we follow the input window length and overlap settings in TSF~\citep{zhang2026tsf}. For WEAR, an acceleration-only dataset incompatible with the gyroscope-dependent IF-ConvTransformer and TSF, we use the 2-second window and 1-second overlap proposed in the original dataset paper. Dataset and preprocessing details are provided in Appendix~\ref{sec:appendix-har-setup}. For the FM baselines and \Name{}, we report the result of 2-layer MLP probe. In multi-sensor datasets, we extract feature from each sensor separately using the FM, and concatenate the features before the MLP. For the supervised baselines, we follow the protocols in TSF. We adopt leave-one-subject-out cross validation (LOSO CV) for all evaluation and report the macro F1 scores (mF1) in Table~\ref{tab:exp-conventional-har}.

\begin{table}[htbp]
\scriptsize
\centering
\setlength{\tabcolsep}{5pt}
\begin{tabular}{llcccccccc}
\toprule
Type                                                                                    & Method       & MoSense        & HAPT           & HHAR           & DSADS          & WEAR           & OPPO.          & Shoaib         & Average        \\ \midrule
\multirow{4}{*}{\begin{tabular}[c]{@{}l@{}}Supervised\\ Baselines\end{tabular}}         & DeepConvLSTM & 93.39          & 82.00          & 91.90          & 83.25          & 75.43          & 40.88          & 97.13          & 80.57          \\
                                                                                        & A\&D         & 95.79          & 81.95          & 94.17          & 86.12          & 77.10          & 46.29          & 97.79          & 82.74          \\
                                                                                        & IF-ConvT.    & 95.81          & 85.35          & 95.01          & 88.82          & N/A            & 50.29          & 98.28          & 85.59          \\
                                                                                        & TSF          & 97.34          & 85.89          & {\ul 95.81}    & 89.70          & N/A            & 52.44          & 99.35          & 86.76          \\ \midrule
\multirow{4}{*}{\begin{tabular}[c]{@{}l@{}}Self-supervised\\ HAR FMs\end{tabular}}      & SimCLR$^*$   & 96.08          & 80.56          & 87.56          & 91.56          & 88.48          & 43.50          & 99.29          & 83.86          \\
                                                                                        & Yuan et al.  & 93.41          & 64.66          & 80.41          & 83.37          & 83.10          & 18.28          & 97.30          & 74.36          \\
                                                                                        & RelCon       & 81.93          & 55.75          & 70.57          & 81.04          & 81.55          & 19.80          & 97.32          & 69.64          \\
                                                                                        & HAR-FM       & 84.93          & 71.41          & 76.47          & 84.93          & 80.01          & 20.24          & 97.67          & 73.20          \\ \midrule
\multirow{2}{*}{\begin{tabular}[c]{@{}l@{}}General-purpose\\ TSFMs\end{tabular}}        & TimesFM 3.0  & 85.64          & 63.30          & 82.97          & 75.09          & 78.92          & 33.50          & 93.61          & 73.29          \\
                                                                                        & Chronos-2    & 83.95          & 57.04          & 72.60          & 73.04          & 62.93          & 24.65          & 90.71          & 66.36          \\ \midrule
\multirow{2}{*}{\begin{tabular}[c]{@{}l@{}}\Name{}        \\ (Ours)\end{tabular}}       & IMUDINO-6M   & {\ul 97.91}    & {\ul 87.56}    & 95.33          & {\ul 93.24}    & {\ul 91.28}    & {\ul 53.54}    & {\ul 99.58}    & {\ul 88.35}    \\
                                                                                        & IMUDINO-85M  & \textbf{98.54} & \textbf{89.02} & \textbf{96.00} & \textbf{94.13} & \textbf{92.34} & \textbf{58.40} & \textbf{99.68} & \textbf{89.73} \\ \bottomrule
\end{tabular}
\caption{Results on seven HAR datasets. \Name{} consistently outperforms both supervised and SSL baselines. $^*$Implemented with \Name{} data augmentation.}
\label{tab:exp-conventional-har}
\end{table}

The results show that \Name{} can be used off-the-shelf for HAR and achieve state-of-the-art performance with lightweight classification heads. Since these datasets cover a wide range of sampling rates, input lengths and sensor locations, this also demonstrates \Name{}'s ability to adapt to diverse scenarios without finetuning. Visualization of \Name{} global features is in Appendix~\ref{sec:appendix-feat-visualization}. We further explore the potential of \Name{}'s global features in complex HAR systems through an online open-set HAR experiment in Appendix~\ref{sec:appendix-activityseeker}. Finally, we evaluate \Name{} on four health and gait analysis tasks (Appendix~\ref{sec:exp-health}), where it also achieved competitive performance.

\subsection{Inertial MoCap: \Name{} Provides Fine-grained Dense Features}

\begin{figure*}[htbp]
  \centering
  \includegraphics[width=0.9\textwidth]{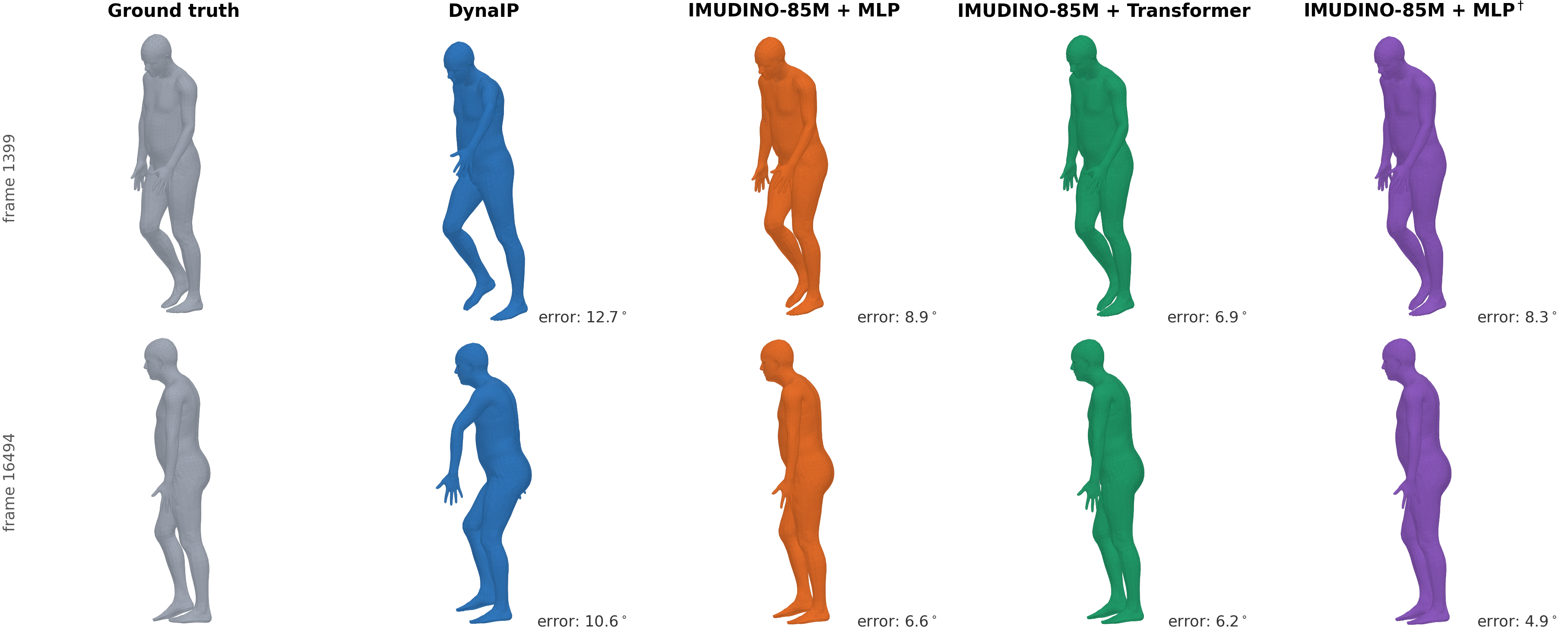}
  \caption{Qualitative comparison of inertial HPE on a natural motion sequence from Virginia Natural Motion. From left to right: ground truth, DynaIP, and the three \Name{} variants of Table~\ref{tab:exp-mocap} (MLP probe, transformer probe, and MLP probe with finetuned backbone). Numbers below each mesh denote the per-frame mean root-aligned global rotation error ($^\circ$). See Appendix~\ref{sec:appendix-mocap-visualization} for more cases.}
  \label{fig:exp-mocap}
\end{figure*}

Motion capture (MoCap) is an important application of inertial sensing. Inertial human pose estimation (HPE) is a challenging task, and existing work often include physical priors in the model and training objective. We explore whether \Name{} can achieve competitive HPE performance \textit{without} these priors. Since MoCap datasets often omit raw gyroscope readings, we synthesize the gyro channels as the finite difference of the $SO(3)$ sequence (Appendix~\ref{app:mocap-protocol}) used by DynaIP. In our experiments, we simply concatenate the patch tokens extracted from each sensor along the channel dimension without strong priors about the human body. Note that the input includes a head-mounted IMU, a location unseen in our training set. We then train the decoder, either an MLP or a transformer, to decode the root-relative 6D pose of 16 joints from the patch tokens using plain MSE loss. We use the AnDy~\citep{maurice2019andy}, UNIPD~\citep{guidolin2022unipd}, CIP~\citep{palermo2022cip}, and Virginia Natural Motion~\citep{geissinger2020virginianaturalmotion} datasets and follow DynaIP's Xsens-only experiment protocols~\citep{zhang2024dynamic}. Detailed experiment settings are in Appendix~\ref{app:mocap-protocol}. We report the results in Table~\ref{tab:exp-mocap} and a visualized example in Fig.~\ref{fig:exp-mocap}.

\begin{table*}[htbp]
\centering
\scriptsize
\setlength{\tabcolsep}{2.5pt}
\begin{tabular}{l|ccc|ccc|ccc|ccc|ccc}
\toprule
\multirow{2}{*}{Model}
& \multicolumn{3}{c|}{AnDy}
& \multicolumn{3}{c|}{UNIPD}
& \multicolumn{3}{c|}{CIP}
& \multicolumn{3}{c|}{Natural Motion}
& \multicolumn{3}{c}{Average} \\
& SIP$\downarrow$ & Ang$\downarrow$ & Pos$\downarrow$
& SIP$\downarrow$ & Ang$\downarrow$ & Pos$\downarrow$
& SIP$\downarrow$ & Ang$\downarrow$ & Pos$\downarrow$
& SIP$\downarrow$ & Ang$\downarrow$ & Pos$\downarrow$
& SIP$\downarrow$ & Ang$\downarrow$ & Pos$\downarrow$ \\
\midrule
DIP~\citep{huang2018dip}
& 11.39 & 5.73 & 4.34
& 12.02 & 5.47 & 4.22
& 19.13 & 8.61 & 6.86
& 33.43 & 11.91 & 13.33
& 18.99 & 7.93 & 7.19 \\

TransPose~\citep{yi2021transpose}
& 12.15 & 6.29 & 4.91
& 15.11 & 6.05 & 4.82
& 20.06 & 8.75 & 6.86
& 30.62 & 11.26 & 11.99
& 19.49 & 8.09 & 7.15 \\

TIP~\citep{jiang2022tip}
& 10.11 & 4.55 & 3.56
& 9.85 & 4.06 & 2.78
& 13.05 & 5.67 & 4.30
& 22.06 & 7.90 & 7.92
& 13.77 & 5.55 & 4.64 \\

PIP~\citep{yi2022pip}
& {\ul 9.49} & 4.09 & \textbf{3.29}
& {\ul 8.90} & 3.59 & 2.66
& 12.68 & 5.52 & 4.12
& 19.62 & 7.49 & 6.93
& 12.67 & 5.17 & 4.25 \\

DynaIP~\citep{zhang2024dynamic}
& \textbf{8.93} & \textbf{3.45} & {\ul 3.41}
& \textbf{7.29} & \textbf{2.77} & \textbf{2.21}
& \textbf{11.42} & \textbf{4.54} & \textbf{3.69}
& \textbf{15.78} & {\ul 7.18} & \textbf{5.83}
& \textbf{10.86} & \textbf{4.49} & \textbf{3.79} \\
\midrule
\Name-85M+MLP
& 11.64 & 4.24 & 3.99
& 11.66 & 3.95 & 3.29
& 16.25 & 5.87 & 5.13
& 19.03 & 7.30 & 6.44
& 14.65 & 5.34 & 4.71 \\

\Name-85M+Transformer
& 10.83 & 3.88 & 3.50
& 10.82 & 3.69 & 2.92
& 14.04 & 5.25 & 4.39
& 19.54 & 7.28 & 6.61
& 13.81 & 5.02 & 4.36 \\

\Name-85M+MLP$^\dagger$
& 10.15 & {\ul 3.72} & 3.43
& 9.45 & {\ul 3.38} & {\ul 2.50}
& {\ul 12.19} & {\ul 4.76} & {\ul 3.95}
& {\ul 17.35} & \textbf{6.65} & {\ul 5.92}
& {\ul 12.28} & {\ul 4.63} & {\ul 3.95} \\
\bottomrule
\end{tabular}
\caption{Inertial MoCap results. We report SIP error ($^\circ$), global angular error ($^\circ$), and joint position error (cm). Average denotes the macro-average over AnDy, UNIPD, CIP, and Natural Motion. $^\dagger$Backbone initialized from the pretrained \Name{} checkpoint and finetuned.}
\label{tab:exp-mocap}
\end{table*}

The frozen MLP probe had TIP-level performance, while the MLP probe with a finetuned backbone performed better than PIP, trailing only DynaIP. Considering that \Name{} is not aided by physical priors and does not have access to the raw gyroscope readings (only reconstructed ones), this result, despite not beating the current SOTA (DynaIP), still shows that \Name{}'s dense features contain rich information about human motion that can be used for pose estimation. With carefully crafted, physics-informed decoder architectures and training objectives, \Name{} can plausibly become a serious contender in inertial HPE. More visualized HPE results are in Appendix~\ref{sec:appendix-mocap-visualization}.

\subsection{Robot Odometry: \Name{} Generalizes Across Embodiments}

Here, we compare \Name{} to TartanIMU~\citep{zhao2025tartanimu}, a robot odometry FM. This task is entirely OOD for \Name{} since it is trained exclusively on wearable IMU data and has no exposure to robot IMUs, and it is kept frozen in this experiment. We train MLP and transformer probes on the 2026 TartanIMU Challenge\footnote{https://superodometry.com/imuchallenge/} train split, and report the results on the official validation split (the test split labels are yet to be released at the time of writing) in Table~\ref{tab:exp-tartan}. The evaluation covers various robot platforms, including car, dog, drone, and human~\citep{zhao2025tartanimu}. TartanIMU has an expert model for each platform, and we report both platform-agnostic performance (prediction averaged over 4 experts) and the results of the specialized experts. \Name{} is evaluated under the \textit{platform-agnostic} setting. We report the standard AVE and ATE20 metrics, and the TartanIMU Challenge score. Visualized examples are in Fig.~\ref{fig:exp-tartan}. Experiment settings are in Appendix~\ref{app:tartan-protocol}.

\begin{table}[htbp]
\centering
\scriptsize
\setlength{\tabcolsep}{3pt}
\begin{tabular}{l|cc|cc|cc|cc|ccc}
\toprule
\multirow{2}{*}{Model}  & \multicolumn{2}{c|}{Car}            & \multicolumn{2}{c|}{Dog}            & \multicolumn{2}{c|}{Drone}          & \multicolumn{2}{c|}{Human}          & \multicolumn{3}{c}{Average}                       \\
                        & AVE$\downarrow$ & ATE20$\downarrow$ & AVE$\downarrow$ & ATE20$\downarrow$ & AVE$\downarrow$ & ATE20$\downarrow$ & AVE$\downarrow$ & ATE20$\downarrow$ & AVE$\downarrow$ & ATE20$\downarrow$ & Score$\downarrow$ \\ \midrule
TartanIMU$^*$              & 0.852           & 2.273             & 0.425           & 1.901             & 1.374           & 2.772             & 0.235           & 1.877             & 0.721          & 2.206            & 0.872            \\
TartanIMU$^\dagger$      & {\ul 0.199}     & 0.999             & 0.129           & {\ul 0.644}       & 0.966     & 2.006             & {\ul 0.107}     & \textbf{0.814}    & 0.350          & 1.116            & 0.429            \\
IMUDINO-85M+MLP$^*$        & 0.200           & {\ul 0.998}       & {\ul 0.123}     & 0.707             & {\ul 0.543}           & {\ul 1.304}       & 0.132           & 1.039             & {\ul 0.249}    & {\ul 1.012}      & {\ul 0.333}      \\
IMUDINO-85M+Transformer$^*$ & \textbf{0.141}  & \textbf{0.760}    & \textbf{0.092}  & \textbf{0.501}    & \textbf{0.499}  & \textbf{1.285}    & \textbf{0.105}  & {\ul 0.872}       & \textbf{0.209} & \textbf{0.854}   & \textbf{0.280}   \\ \bottomrule
\end{tabular}
\caption{Results on the TartanIMU Challenge. $^*$Platform-agnostic. $^\dagger$Platform expert.}
\label{tab:exp-tartan}
\end{table}

The results show that \Name{} outperforms the existing SOTA, TartanIMU, on robot inertial odometry despite the task being OOD. This frozen-backbone cross-embodiment transfer indicates that \Name{} dense features can readily transfer to unseen dynamics, in this case robot motion. To our knowledge, \Name{} is the first inertial FM to demonstrate such capabilities, making progress towards \textit{general-purpose foundation models} for the broader inertial sensing community.

\begin{figure*}[htbp]
  \centering
  \includegraphics[width=0.8\textwidth]{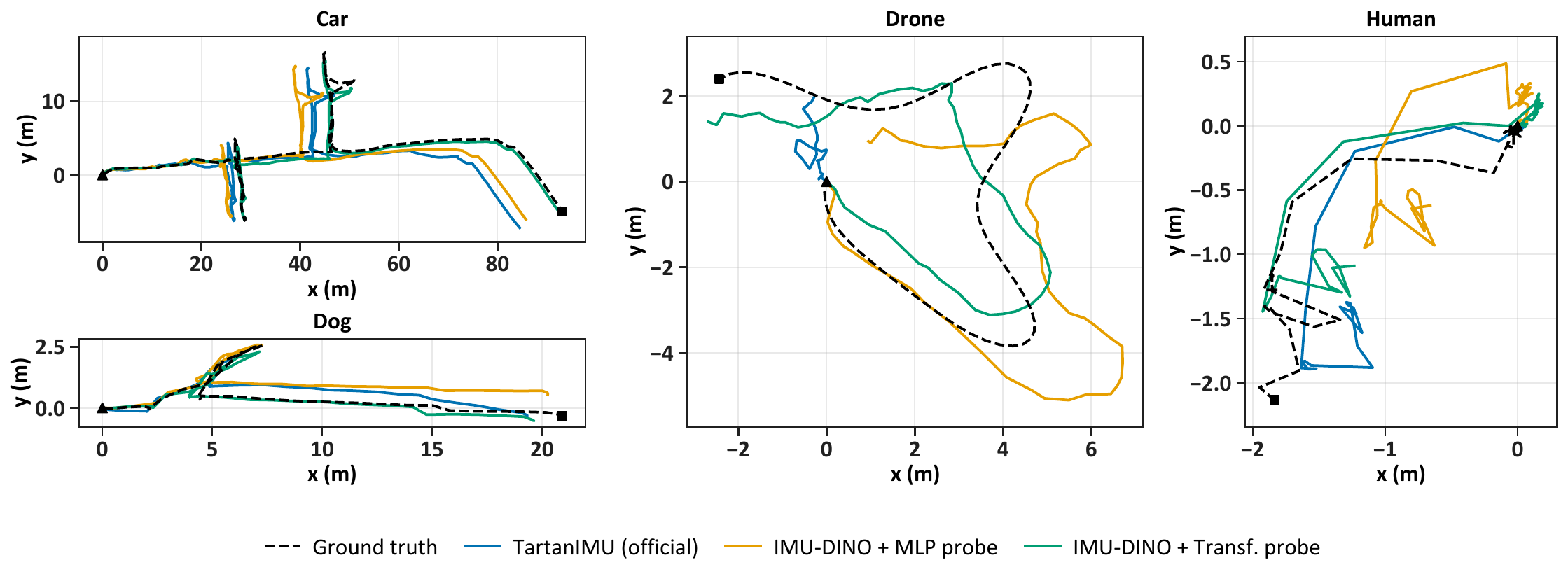}
  \caption{Visualized results on the TartanIMU Challenge. $\blacktriangle$ denotes the starting point of each trajectory. \Name{}+Transformer (green line) traces the ground truth (black dashed line) most closely. The TartanIMU results (blue line) are from the corresponding expert model for each platform.}
  \label{fig:exp-tartan}
\end{figure*}

\subsection{Ablation Study}

In our ablation study, we dissect how each component of \Name{}'s training recipe contribute to overall performance. For training objective, we consider the impact of the patch-level objective and the KoLeo regularizer. For data augmentation, we analyze the effect of scaling, rotation, resampling and channel masking. We also explore the contribution of frequency-aware PE. Finally, we include results of randomly initialized backbones to quantify the gain from SSL pretraining. IMUFormer-85M is used in all experiments. We report HAR, MoCap and robot odometry results in Table~\ref{tab:exp-ablation}.

\begin{table}[htpb]
\centering
\scriptsize
\begin{tabular}{ll|c|ccc|ccc}
\toprule
\multirow{2}{*}{Type}                                                                  & \multirow{2}{*}{Setting}        & HAR            & \multicolumn{3}{c|}{MoCap (MLP Probe)}              & \multicolumn{3}{c}{Robot Odometry (MLP Probe)}          \\
                                                                                       &                                 & mF1            & SIP$\downarrow$ & Ang$\downarrow$ & Pos$\downarrow$ & AVE$\downarrow$ & ATE20$\downarrow$ & Score$\downarrow$ \\ \midrule
\multirow{2}{*}{\begin{tabular}[c]{@{}l@{}}Training\\ Objective\end{tabular}}          & w/o patch-level objective       & 89.41          & 15.21            &  5.77             &  4.85             & 0.293 & 1.045 & 0.373 \\
                                                                                       & w/o KoLeo                       & 89.40          & 14.50            &  5.31             &  4.58             & 0.252 & 1.004 & 0.335 \\ \midrule
\multirow{4}{*}{\begin{tabular}[c]{@{}l@{}}Data\\ Augmentation\end{tabular}}           & w/o scaling                     & 89.63          & 14.87           & 5.35            & 4.74            & 0.257 & 1.016 & 0.340 \\
                                                                                       & w/o rotation                    & 88.66          & 15.67            &  5.52             &  4.93             & 0.253 & 1.036 & 0.339 \\
                                                                                       & w/o resampling                  & 88.86          & 14.64            &  5.25             &  4.59             & 0.255 & 1.012 & 0.338 \\
                                                                                       & w/o channel masking             & 89.04          & 14.02           & 5.19            & 4.41            & 0.258 & 1.023 & 0.342 \\ \midrule
Model                                                                                  & w/o freq-aware PE               & 89.60          & 14.51           & 5.26            & 4.52            & 0.260           & 1.022             & 0.343             \\ \midrule
\multirow{2}{*}{\begin{tabular}[c]{@{}l@{}}Self-supervised\\ Pretraining\end{tabular}} & random init, frozen backbone    & 65.32          & 17.34            &  5.90             &  5.62             & 0.854 & 6.976 & 1.592 \\
                                                                                       & random init, trainable backbone & 83.70          & 14.89            &  5.32             &  4.77             & 0.534 & 2.852 & 0.802             \\ \midrule
IMUDINO-85M                                                                            &                                 & 89.73          & 14.65           & 5.34            & 4.71            & 0.249 & 1.012 & 0.333             \\ \bottomrule
\end{tabular}
\caption{Result of the ablation study. The results are averaged over multiple datasets and correspond to the \textit{Average} columns in Tables~\ref{tab:exp-conventional-har},~\ref{tab:exp-mocap} and~\ref{tab:exp-tartan}.}
\label{tab:exp-ablation}
\end{table}

The ablation of training objectives show that the inclusion of a patch-level objective effectively improved \Name{}'s dense features. While the KoLeo regularizer lead to a slight drop in MoCap performance, it improved HAR performance and made training more stable. For data augmentation, applying scaling and rotation led to improvements across the board. Resampling led to better HAR performance but virtually no change in MoCap and robot odometry performance. Considering resampling is key to \Name{}'s support of variable sampling rate, we kept resampling in the final recipe. Channel masking led to improved HAR performance but reduced MoCap metrics. However, we observed improved training stability with channel masking, so it is kept as well. Frequency-aware PE is positive overall, improving HAR and odometry while slightly hindering MoCap. Finally, \Name{} outperformed a trainable, randomly-initialized backbone in all 3 tasks, indicating that our SSL pretraining yielded gains beyond the reach of conventional supervised learning.

\textbf{More ablation results.} An HAR stress test for variable input length and sampling rate is in Appendix~\ref{sec:appendix-stress-test}. A scaling experiment is in Appendix~\ref{sec:appendix-scaling}.

\subsection{Balancing Global and Dense Representations}

\begin{figure*}[htbp]
  \centering
  \includegraphics[width=0.9\textwidth]{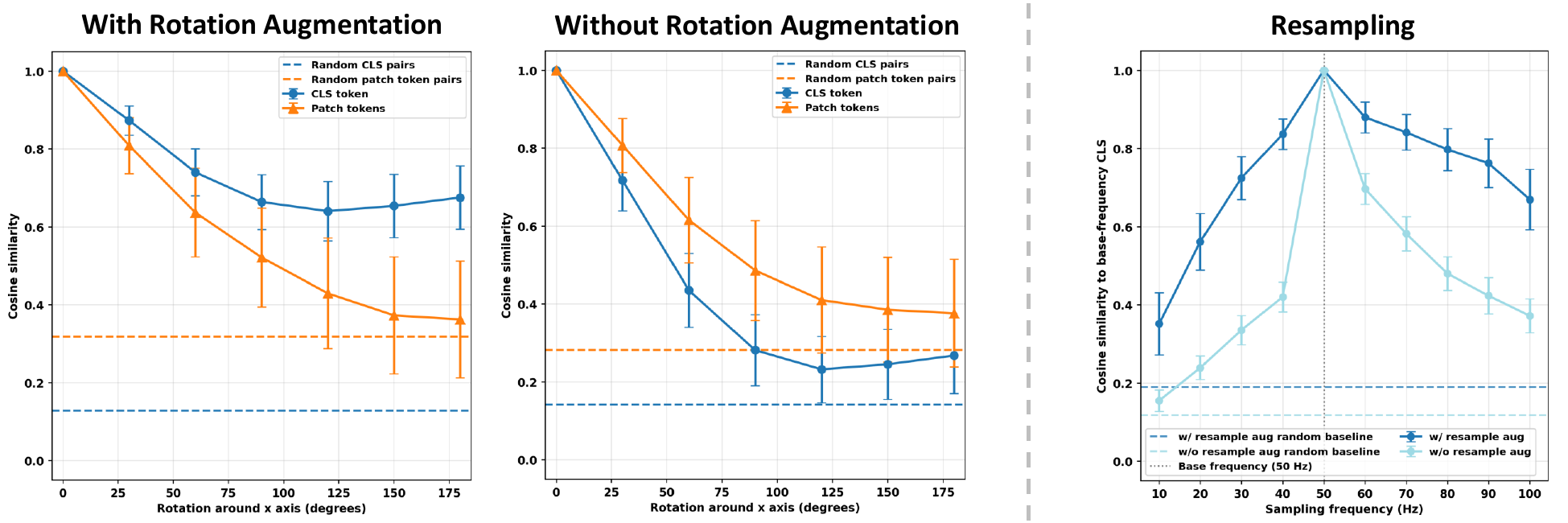}
  \caption{How CLS and patch tokens responds to perturbation. Our invariance-motivated view transformations induced in the CLS tokens the desired robustness to rotation and resampling. We further demonstrate in the rotation case that CLS robustness does not expand to patch tokens, which remain sensitive to rotation. Results aggregated over 700 examples from 7 HAR datasets.}
  \label{fig:discussion-sim}
\end{figure*}

How does \Name{} balance global and patch-level representation? A general-purpose inertial FM needs to produce high-quality global and patch-level representations simultaneously. This creates a natural tension: global features demand semantic invariance, while dense features need to capture local variations. \Name{}'s training objective circumvents this conflict by applying the invariance-inducing view transformations only to the global objective $\mathcal{L}_{\mathrm{DINO}}$, while the patch-level representation remains sensitive to these transformations. In Fig.~\ref{fig:discussion-sim}, we compare the cosine similarity between tokens after different rotations around the x axis and resampling. 

In the model trained with rotation augmentation, the CLS similarity decreased only marginally when the input is rotated. The model trained without rotation sees a much more drastic decrease in similarity. In contrast, the patch token similarity trend is almost identical, decreasing to near-random level when we rotate 180$^\circ$, indicating that orientation is retained in the dense features. In addition, we also demonstrate how the \textit{resampling} transformation improve robustness to varying sampling rate. Again, data augmentation induced the desired invariance in global feature representation. 

\section{Conclusion}

We introduced \Name{}, a foundation model for inertial sensing. We adapted DINO self-distillation to inertial SSL through a set of global semantics-preserving view transformations, as well as a minimalist transformer architecture. We trained \Name{} on a relatively modest corpus of 11600 hours of wearable inertial data covering various scenarios and sensor setups. Extensive experiments on HAR, health, motion capture and robot odometry demonstrated that this simple approach is remarkably effective, and produces generalizable global and dense features. To the best of our knowledge, it is the first inertial FM to demonstrate competitive performance on a broad range of global and dense prediction tasks, including OOD ones like robot odometry. \Name{} demonstrated that it is possible to learn broadly generalizable motion representation from inertial data, and that inertial FMs do not need to narrowly focus on one sub area (e.g., HAR). We hope \Name{} can empower inertial sensing applications and kickstart research into general-purpose inertial FMs.

\bibliography{iclr2027_conference}
\bibliographystyle{iclr2027_conference}

\appendix
\section{Implementation Details}
\label{sec:appendix-impl-details}
\subsection{Motion Transformations}
\label{sec:appendix-impl-details-transform}
We report how each motion transformation is implemented below.

\textbf{Cropping.} We sample the global crop size from $U(0.5, 0.8)$ and the local crop size from $U(0.05, 0.4)$. We then randomly choose the starting point of each crop.

\textbf{Scaling.} Scaling is randomly applied to the views with a probability of $p=0.5$. Both the speed factor and magnitude factor are sampled from $U(0.8, 1.2)$. Speed up / down is implemented with linear interpolation, while the magnitude factor is simply multiplied to the sensor readings. Empirically, mild scaling like this rarely change the global semantics of motion significantly.

\textbf{Rotation.} Rotation is randomly applied to the views with a probability of $p=0.5$. We sample a random rotation matrix from $SO(3)$. This covers the full $-180^\circ$ to $180^\circ$ range on all three axes.

\textbf{Resampling.} Resampling is randomly applied to the views with a probability of $p=0.5$. The new frequency (in Hz) is sampled from $U(30, 200)$. We use linear interpolation to implement resampling.

\textbf{Channel masking.} Channel masking is applied to each channel with a probability of $p=0.2$. The masked channel is replaced with zeros.

\subsection{Models and Training Recipe}
\label{sec:appendix-impl-details-model-training}
\textbf{Models.} Detailed model hyperparameters are in Table~\ref{tab:appendix-model-parameters}. IMUFormer-85M is identical to ViT-B except for the number of attention heads, which we increased from 12 to 24.

\begin{table}[htbp]
\centering
\small
\begin{tabular}{lcccc}
\toprule
                & IMUFormer-2.7M & IMUFormer-6M & IMUFormer-25M & IMUFormer-85M \\ \midrule
Patch size      & 16             & 16           & 16            & 16            \\
Token dim       & 192            & 256          & 512           & 768           \\
Num heads       & 6              & 8            & 16            & 24            \\
Num layers      & 6              & 8            & 8             & 12            \\
Register Tokens & 4              & 4            & 4             & 4             \\
MLP type        & GELU MLP       & GELU MLP     & GELU MLP      & GELU MLP      \\
MLP ratio       & 4              & 4            & 4             & 4             \\
Drop path       & 0.0            & 0.0          & 0.1           & 0.3           \\ \bottomrule
\end{tabular}
\caption{The 4 IMUFormer models used in this paper.}
\label{tab:appendix-model-parameters}
\end{table}

\textbf{Training Recipe.} We train the models on 6 Nvidia H100 GPUs using PyTorch DDP and BF16 AMP. We use a batch size of 1536 samples, leading to 3072 global views and 9216 local views per batch. To improve training efficiency, we limit global views to 1024 frames and local views to 512 frames. Frames exceeding this cutoff, generated when the \textit{resample} transformation upsamples a crop, are discarded. We train the models for 60 epochs using the AdamW optimizer, with a 1000-step linear warmup, a post-warmup LR of 5e-4, a cosine annealing LR schedule and a weight decay of 5e-2. Similar to DINO, we warmup the teacher's softmax temperature linearly from 0.04 to 0.07 in the first 1000 steps, and gradually increase its EMA momentum from 0.996 to 1.0 following a cosine schedule spanning the entire training run. To improve training stability, we clip the gradient norm to 3.0. As in DINOv2, we compute $\mathcal{L}_{\mathrm{KoLeo}}$ separately for each GPU to maximize efficiency. Training takes 11 hours for IMUFormer-85M and 6 hours for IMUFormer-6M.

\section{Pre-training Dataset Processing}
\label{sec:appendix-pretrain-data-proc}

In this section we present details on how we filtered the pre-training dataset. We generally follow this four step process: file-level quality control, unit normalization, cropping into 1024-frame samples, and sample-level quality control.

\textbf{Capture24.} We removed recordings where the sampling rate deviate from the normal 100 Hz as quality control. We further filtered out segments with null or invalid annotation. The remaining data has the unit of acceleration normalized to $\mathrm{m/s^2}$ (default unit is $g$).

\textbf{SHL.} We did not filter SHL data, as all SHL data has consistent sampling rate and no NaN values. We simply combine the train and validation sets from the SHL 2026 Challenge.

\textbf{PAAWS.} First, we removed all data where the ``Sensor orientation issue(s)'' field is not empty or ``None'' (352 files). We also removed data where the sampling rate deviates significantly from the expected 100 Hz (3 files). For the remaining files, we removed all NaN readings, and normalized the unit of acceleration from $g$ to $\mathrm{m/s^2}$ and that of angular velocity from $\mathrm{deg/s}$ to $\mathrm{rad/s}$. Finally, we filtered samples where with inconsistent activity / posture labels, as well as static samples with linear acceleration below 0.30 $\mathrm{m/s^2}$ \textbf{and} angular velocity below 0.08 $\mathrm{rad/s}$.

\section{Visualization of \Name{} Features}
\label{sec:appendix-feat-visualization}
\begin{figure*}[htbp]
  \centering
  \includegraphics[width=1.0\textwidth]{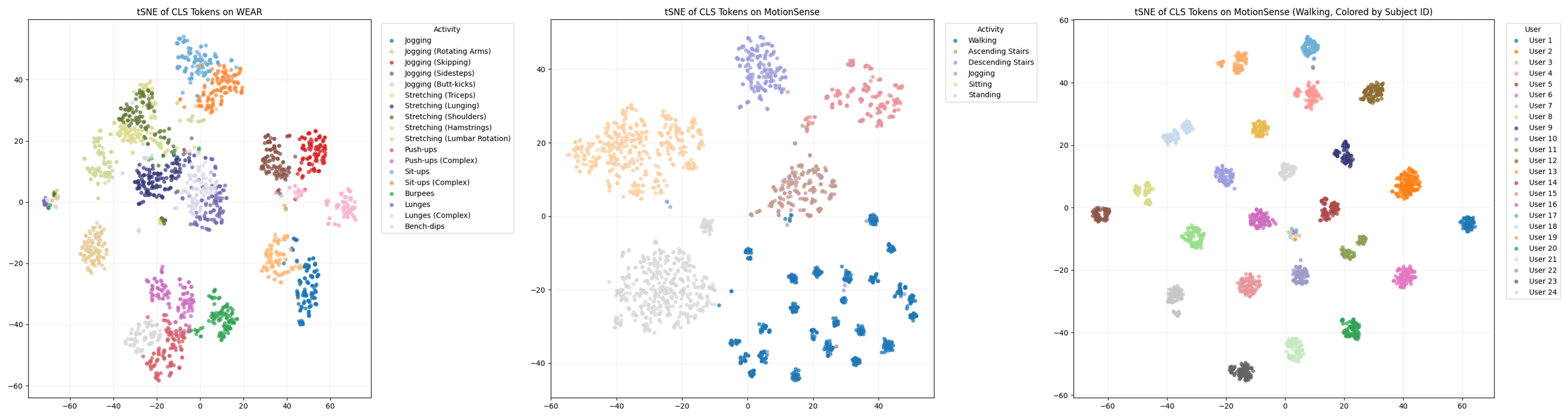}
  \caption{\Name{} feature visualization. Samples are colored by label. Activities from WEAR (left) and MotionSense (middle) form distinct clusters. Furthermore, for the \textit{walking} class in MotionSense (right), each user forms a distinct cluster, indicating that \Name{} is capable of capturing the differences in gait between individuals. Further results on gait recognition are in Appendix~\ref{sec:exp-health}.}
  \label{fig:appendix-feature}
\end{figure*}

We present a t-SNE visualization of \Name{} features in Fig.~\ref{fig:appendix-feature}. \Name{} features can distinguish different activities, as well as different people performing the same activity.

\section{HAR Experiment Setup}
\label{sec:appendix-har-setup}
\subsection{Datasets}
Seven datasets are used in our HAR experiment: MotionSense~\citep{malekzadeh2018motionsense}, HAPT~\citep{reyes-ortiz2016hapt}, HHAR~\citep{stisen2015hhar}, DSADS~\citep{barshan2014dsads}, OPPORTUNITY~\citep{chavarriaga2013opportunity}, \citet{shoaib2014shoaib}, and WEAR~\citep{bock2024wear}. We provide a detailed description of the number of subjects, activity classes, and sensor setups of the datasets, as well as the protocols used in our experiments.

\textbf{MotionSense} is a smartphone IMU dataset covering 24 subjects and 6 activities (ascending stairs, descending stairs, walking, jogging, sitting, and standing). The smartphone was placed in the subjects' front pockets, and the experiment was done in a controlled setting. The sampling rate is 50 Hz.

\textbf{HAPT} is an extended version of the classic UCI-HAR~\citep{anguita2013ucihar} smartphone IMU dataset covering 30 subjects and 12 activities. These include 6 normal activities (ascending stairs, descending stairs, walking, sitting, standing, and lying down), and 6 types of transitions (stand-to-sit, sit-to-stand, sit-to-lie, lie-to-sit, stand-to-lie, and lie-to-stand). The smartphone was waist-mounted and had a sampling rate of 50 Hz.

\textbf{HHAR} is a smartphone dataset covering 9 subjects and 6 activities (ascending stairs, descending stairs, walking, cycling, sitting, and standing). Each subject carried 8 smartphones (4 different models) in a tight pouch secured around their waist. The sampling rates ranged from 50 Hz to 200 Hz, depending on the phone model.

\textbf{DSADS} is a wearable IMU dataset covering 8 subjects and 19 activities (walking in a parking lot, 4 km/h walking on flat treadmill, 4 km/h walking on inclined treadmill, 8 km/h running on treadmill, ascending stairs, descending stairs, exercising on a stepper, exercising on a cross trainers, cycling on an exercise bike in horizontal position, cycling on an exercise bike in vertical position, rowing, jumping, playing basketball, sitting, standing, lying on back, lying on right side, standing still in an elevator, and moving around in an elevator). 5 IMUs were placed on each subject (torso, right arm, left arm, right leg, and left leg). The IMUs had a sampling rate of 25 Hz.

\textbf{OPPORTUNITY} is a wearable IMU dataset covering 4 subjects in a sensor rich environment. Following \citet{zhang2026tsf}, we use the most challenging 18-class setting (null class, open door 1, open door 2, close door 1, close door 2, open fridge, close fridge, open dishwasher, close dishwasher, open drawer 1, open drawer 2, open drawer 3, close drawer 1, close drawer 2, close drawer 3, clean table, drink from cup, and toggle switch). We use the seven 6-axis IMUs (back, right upper arm, left upper arm, right lower arm, left lower arm, left shoe, right shoe) as the input. The data was collected at 30 Hz.

\textbf{WEAR} is a wearable IMU dataset covering 22 subjects and 18 workout activities, including 5 types of jogging, 5 types of stretching, and 8 types of strength training. It also included a null class. The data is collected from four 50 Hz accelerometers (one on each ankle and one on each wrist).

\textbf{\citet{shoaib2014shoaib}}'s HAR dataset covers 10 subjects and 7 activities ((ascending stairs, descending stairs, walking, jogging, cycling, sitting, and standing). The data is collected from 5 smartphones (right jeans pocket, left jeans pocket, belt clipper, right upper arm, and right wrist) at 50 Hz.

Combined, these datasets cover a huge variety of sensor locations, including on the limbs and torso, as well as in pockets and pouches. \Name{}'s SOTA performance across all seven datasets indicates strong generalization across sensor locations.

\subsection{Protocol}

We train a 2-layer MLP probe on top of frozen \Name{} features (or frozen features from the FM baselines). The MLP has the following structure: Dropout(0.2)~$\to$~Linear($D\times S \to 512$)~$\to$~GELU~$\to$~Dropout(0.5)~$\to$~Linear($512 \to K$), where $D$ is the feature dimension of the backbone, $S$ is the number of sensors (features are concatenated along the \textit{channel} dimension) and $K$ the number of classes. We use a mild data augmentation scheme: random scaling ($p=0.5$) identical to \Name{} pretraining settings, random rotation ($p=0.5$) along the three axes sampled from $U(-30^\circ,30^\circ)$, and random jitter ($p=0.5$, Gaussian noise from $\mathcal{N}(0,0.05)$). We use the AdamW optimizer with a weight decay of 0.1 and a cosine LR schedule. Detailed data processing protocols, and the LR for each dataset, are in Table~\ref{tab:har-protocol}. All experiments follow the leave-one-subject-out (LOSO) cross validation protocol. The sampling rate, window size, window overlap, and LOSO protocol we used are identical to those of TSF~\citep{zhang2026tsf}, making the HAR results directly comparable except for WEAR, an accelerometer-only dataset lacking the gyroscope channels required by TSF. For WEAR, we follow the 2-second window with 1-second overlap setup in the original dataset paper~\citep{bock2024wear}.

\begin{table}[ht]
\centering
\scriptsize
\label{tab:har_datasets}
\begin{tabular}{|l|c|c|c|c|c|c|c|}
\hline
\textbf{Datasets} 
& MotionSense 
& HAPT 
& HHAR 
& DSADS 
& WEAR 
& OPPORTUNITY
& \citet{shoaib2014shoaib} \\ 
\hline

Sampling Rate (Hz) 
& 50 
& 50 
& 100$^*$
& 25 
& 50 
& 30 
& 50 \\ 
\hline

Window Size (s)
& 2.56 
& 2.56
& 2.00
& 5.00 
& 2.00
& 1.60
& 0.96 \\ 
\hline

Overlap (s)
& 1.28 
& 1.28
& 1.00 
& 0.00 
& 1.00
& 0.80 
& 0.48 \\ 
\hline

LR 
& 1e-3
& 1e-3
& 1e-3
& 1e-4
& 3e-4
& 3e-4
& 3e-4 \\ 
\hline

Epochs 
& 100
& 100
& 100
& 100
& 50
& 50
& 50 \\ 
\hline

\end{tabular}
\caption{Data processing protocol, LR, and the number of training epochs for each HAR dataset. We use the final epoch checkpoint for test set evaluation. $^*$Phones with different sampling rate are resampled to 100 Hz.}
\label{tab:har-protocol}
\end{table}

\section{Online Open-set HAR}
\label{sec:appendix-activityseeker}

In addition to conventional HAR, we consider the challenge of \textit{personalization}. Online personalized HAR aims to solve the problem by learning user-specific patterns through actual use and interactions. One example is ActivitySeeker~\citep{ye2026activityseeker}, a recent personalized HAR system that (1) extracts features from unlabeled IMU data, (2) discovers motion patterns through clustering, (3) asks the user for a label when the current activity matches a cluster, (4) labels the entire cluster with the user-provided information, and (5) trains lightweight classifiers using the personalized dataset. The performance of ActivitySeeker's \textit{collaborative labeling} strategy is highly dependent on the feature extraction model. Moreover, the ActivitySeeker dataset covers diverse sensor locations and activities. These factors make ActivitySeeker a suitable test bench for \Name{}.

In this experiment, we reproduce the simulated online learning experiment of~\citet{ye2026activityseeker} and replace the original ResNet feature extraction model with \Name{}. Note that \Name{}-6M is around the same size as the ResNet in the original paper. We report the metrics used by~\citet{ye2026activityseeker} in Table~\ref{tab:exp-activityseeker}. All metrics are averaged over 13 users. The results demonstrate that motion patterns are revealed much more efficiently when \Name{} features are used, leading to higher data utilization and better HAR performance.

\begin{table}[htbp]
\centering
\scriptsize
\begin{tabular}{lcccc}
\toprule
Backbone / Condition         & Data Util.    & Collab. Labeling Acc. & Activities Discovered & HAR mF1          \\
\midrule
W/o Personalization          & -             & -                     & -                     & 62.6             \\
\midrule
Supervised ResNet            & 67.4          & {\ul 98.8}            & 8.3 / 8.7             & 89.5             \\
IMUDINO-6M                   & {\ul 76.5}    & 98.7                  & \textbf{8.4 / 8.7}    & {\ul 95.3}       \\
IMUDINO-85M                  & \textbf{77.3} & \textbf{99.3}         & \textbf{8.4 / 8.7}    & \textbf{96.8}    \\
\bottomrule
\end{tabular}
\caption{Results on ActivitySeeker. Using \Name{} off-the-shelf improves all metrics, demonstrating its potential as a general-purpose inertial feature extraction model.}
\label{tab:exp-activityseeker}
\end{table}

\section{Gait Analysis and Health}
\label{sec:exp-health}

\subsection{Task Overview and Results}

We evaluate \Name{} on a variety of gait analysis and health tasks. These include activity intensity estimation on GOTOV~\citep{paraschiakos2020gotov}, fall detection on UniMiB-SHAR~\citep{micucci2017unimibshar}, freezing of gait (FOG) detection in Parkinson's patients on Daphnet FOG~\citep{roggen2010daphnetfog}, and gait recognition on whuGAIT~\citep{zou2020whugait}. We follow the experiment settings described in the original dataset papers and report the results in Table~\ref{tab:exp-health}. Overall, \Name{} delivers competitive performance on all 4 datasets, demonstrating its potential in diverse health and gait analysis applications.

\begin{table}[htbp]
\centering
\scriptsize
\begin{tabular}{l|cc|cc|cc|cc}
\toprule
\multirow{2}{*}{Method} & \multicolumn{2}{c|}{GOTOV}       & \multicolumn{2}{c|}{UniMiB-SHAR} & \multicolumn{2}{c|}{Daphnet FOG} & \multicolumn{2}{c}{whuGAIT}      \\
                        & RMSE$\downarrow$ & R$^2$         & HAR Acc         & Fall Acc       & Spec.           & Sens.          & Acc            & EER$\downarrow$ \\ \midrule
Official Baseline       & 1.35             & 0.45          & 56.58           & 97.05          & 0.869           & \textbf{0.781} & 93.75          & -               \\
IMUDINO-6M+MLP          & {\ul 0.96}       & {\ul 0.58}    & {\ul 83.96}     & {\ul 99.59}    & {\ul 0.908}     & {\ul 0.671}    & {\ul 96.55}    & {\ul 3.24}      \\
IMUDINO-85M+MLP         & \textbf{0.94}    & \textbf{0.59} & \textbf{84.78}  & \textbf{99.81} & \textbf{0.936}  & 0.551          & \textbf{97.41} & \textbf{2.92}   \\ \bottomrule
\end{tabular}
\caption{Results on four health and gait analysis datasets. Official baseline results are from the original dataset paper, whose experiment protocols are followed in \Name{} experiments.}
\label{tab:exp-health}
\end{table}

\subsection{Common setup}
\label{app:health-common}

\paragraph{Input convention.} Each sensor stream is a 6-channel window:
accelerometer $xyz$ in m/s$^2$ followed by gyroscope $xyz$ in rad/s. For
acceleration-only datasets, the gyroscope channels are zero-filled, matching
the acceleration-only streams seen in pretraining. The per-sample sampling rate
is passed to the frequency-aware positional encoding, so each dataset keeps its
native sampling rate unless stated otherwise.

\paragraph{Probe and optimization.} Unless stated otherwise in the next section, the probe is a
2-layer MLP over the frozen class token(s) of the last layer:
Dropout(0.2)~$\to$~Linear($D\times S \to 512$)~$\to$~GELU~$\to$~Dropout(0.5)~$\to$~Linear($512 \to K$),
where $D$ is the feature dimension of the backbone, $S$ is the number of sensor streams (features are concatenated across
streams) and $K$ the number of classes (1 for regression). A fresh probe is
trained per fold with AdamW (weight decay 0.1), cosine learning-rate decay,
bf16 autocast, and cross-entropy with label smoothing 0.2 (classification) or
MSE on fold-standardized targets (regression). Training windows receive the
same light augmentation as our HAR evaluation (random scaling, $\pm 30^\circ$
rotation, speed and jitter perturbations) unless stated otherwise. Subject
splits follow each dataset's canonical protocol, defaulting to
leave-one-subject-out cross validation (LOSO CV).

\subsection{Datasets and per-task settings}
We provide a summary of the health and gait analysis task settings in Table~\ref{tab:app-health-summary}.

\begin{table}[htbp]
\centering
\scriptsize
\setlength{\tabcolsep}{4pt}
\begin{tabular}{lll}
\toprule
Dataset & Task & Protocol \\
\midrule
whuGAIT & gait authentication & fixed split (20 unseen subjects)\\
UniMiB-SHAR & 17-way HAR / fall detection & LOSO \\
Daphnet & freeze of gait (FOG) detection & LOSO, 4~s/0.5~s windows \\
GOTOV & energy expenditure (EE) regression & LOSO, 120~s$\to$10~s bins \\
\bottomrule
\end{tabular}
\caption{Summary of the gait analysis and health evaluation. All experiments use
frozen \Name{} backbones with lightweight MLP probes.}
\label{tab:app-health-summary}
\end{table}

\paragraph{Energy expenditure (GOTOV~\citep{paraschiakos2020gotov}).}
GOTOV provides synchronized accelerometry (three GENEActiv units at ankle,
wrist and chest, resampled from 85.71~Hz to 50~Hz) and breath-by-breath EE from
a COSMED K4b2 (kcal/min) for 31 older adults. We follow the protocol
of~\citet{paraschiakos2020gotov}: EE labels are aggregated into 10~s bins, and
each bin is predicted from the preceding 120~s acceleration window (10,875
windows, labels 0.70--14.18 kcal/min), evaluated with LOSO CV. The probe is a
$2304 \to 512 \to 1$ MLP (three concatenated sensor features) trained with MSE
on fold-standardized targets, lr $10^{-3}$, 150 epochs, batch 256.

\paragraph{Activity and fall recognition (UniMiB-SHAR~\citep{micucci2017unimibshar}).}
UniMiB-SHAR contains 30 subjects performing 9 activities of daily living (ADL)
and 8 fall types, recorded by a 50~Hz smartphone accelerometer and segmented
into peak-centered 3~s windows (151 samples; truncated to 144 to align with the
patch size). We run 17-way classification with LOSO CV (the dataset's own
protocol), lr $10^{-3}$, 100 epochs, batch 1024.

\paragraph{Freezing-of-gait detection (Daphnet FOG~\citep{roggen2010daphnetfog}).}
Daphnet records 10 Parkinson's patients with three 64~Hz accelerometers
(ankle, thigh, trunk; values converted from mg to m/s$^2$). We follow the
canonical protocol of~\citet{roggen2010daphnetfog}: 4~s windows with 0.5~s step,
a window is labeled \emph{freeze} if at least 50\% of its samples are annotated
as freeze, and segments outside the protocol are discarded, giving 35,405
windows (9.6\% freeze) evaluated with LOSO CV. We use balanced oversampling to mitigate the data imbalance.

\paragraph{Gait authentication (whuGAIT~\citep{zou2020whugait}, Dataset \#5).}
whuGAIT Dataset \#5 contains 74,142 authentication pairs from 118 subjects (66,542
training pairs from 98 subjects; 7,600 test pairs from 20 disjoint subjects),
each pair consisting of two horizontally aligned 128-sample two-step segments
of 50~Hz smartphone accelerometer+gyroscope data. The task is binary
same/different-subject verification on the fixed split. The input here are the concatenated interaction features
$[a, b, |a-b|, a \cdot b]$ over per-segment features (layer-11 class token
concatenated with mean-pooled patch tokens). We also use swap augmentation
$(a,b) \to (b,a)$, an lr of $2.5 \times 10^{-4}$ with 60 epochs and early stopping on
a 10\% validation split carved from the \emph{training} pairs only.

\section{HAR with Variable Input Length and Sampling Rate}
\label{sec:appendix-stress-test}
In this section, we report MLP-probe results on MotionSense under different input length and sampling rate conditions sampled from the following grid: $\mathrm{\{1.28s, 2.56s, 5.12s, 10.24s\}} \times \mathrm{\{25Hz, 50Hz, 100Hz, 200Hz\}}$, in Table~\ref{tab:appendix-motionsense-window-freq}. The \Name{}-85M variant is used. The results show that changing the sampling rate does not impact HAR performance by much, especially for larger windows (5.12s and 10.24s). Likewise, \Name{} handles all 4 window sizes well, with larger windows bringing slight improvements as expected. In summary, this experiment demonstrates that \Name{} does indeed support variable input length and sampling rate.

\begin{table}[htbp]
\centering
\small
\begin{tabular}{lcccc}
\toprule
Window & 25 Hz & 50 Hz & 100 Hz & 200 Hz \\
\midrule
1.28 s  & 97.40 & 97.56 & 97.60 & 97.86 \\
2.56 s  & 98.26 & 98.54 & 98.67 & 98.72 \\
5.12 s & 98.89 & 98.91 & 98.92 & 98.81 \\
10.24 s & 99.16 & 99.34 & 99.33 & 99.24 \\
\bottomrule
\end{tabular}
\caption{MotionSense mF1 of \Name{}-85M with the 2-layer MLP probe for different input window lengths and sampling rates (resampled from the native 50 Hz). Supervised SOTA: 97.34 (2.56 s, 50 Hz, \citet{zhang2026tsf}).}
\label{tab:appendix-motionsense-window-freq}
\end{table}

\section{Scaling \Name{}}
\label{sec:appendix-scaling}
\begin{figure*}[htbp]
  \centering
  \includegraphics[width=1.0\textwidth]{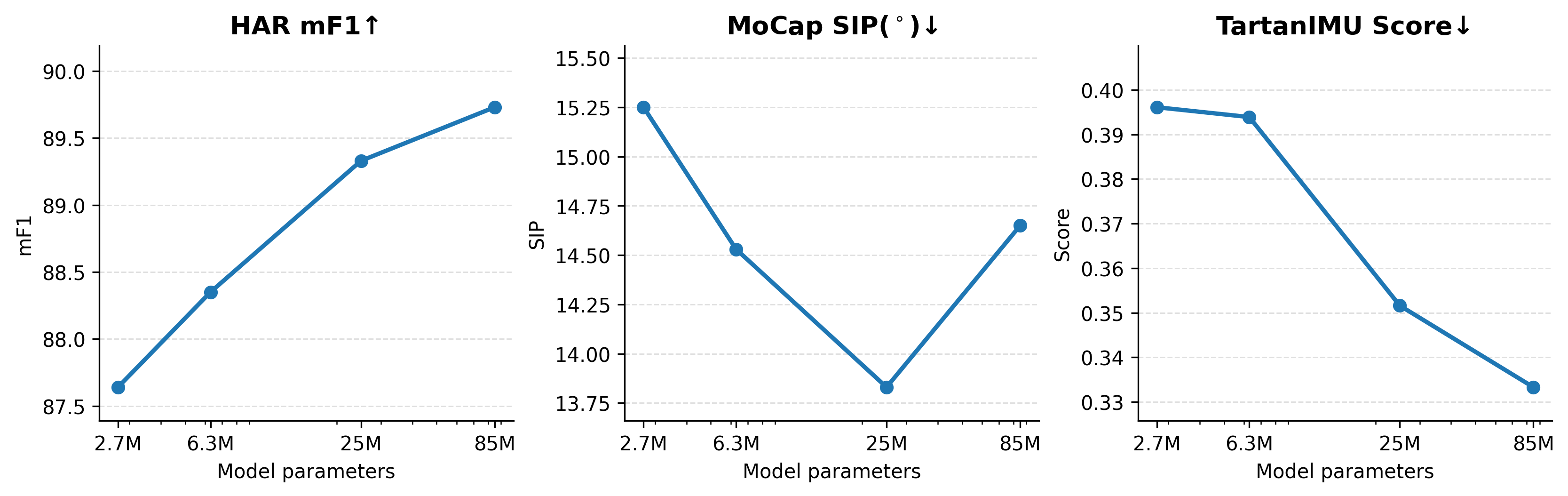}
  \caption{How \Name{} models of different sizes perform in HAR, MoCap, and robot odometry.}
  \label{fig:appendix-scaling}
\end{figure*}

We explore how scaling affects \Name{} performance and report the results in Fig.~\ref{fig:appendix-scaling}. For HAR and robot odometry, larger models consistently perform better than smaller ones. MoCap, however, seems to saturate at 25M parameters. This type of behavior is not uncommon in the DINO paradigm: DINOv2 ViT-L outperformed its larger ViT-g counterpart on instance-level recognition~\citep{oquab2024dinov2}, while DINOv3 ViT-L outperformed its larger ViT-H+ counterpart on segmentation, tracking, and keypoint matching~\citep{simeoni2025dinov3}. Overall, we believe that scaling up \Name{} further is worth exploring as future work as there is still considerable potential in inertial SSL waiting to be unlocked.

\section{MoCap Experiment Setup}
\label{app:mocap-protocol}

\subsection{Task and Datasets}
Given six body-worn IMUs (pelvis, head, left/right lower arm, left/right lower leg; 60\,Hz), the model regresses full-body articulation. Following DynaIP~\citep{zhang2024dynaip}, we use the publicly available Xsens motion-capture datasets (AnDy~\citep{maurice2019andy}, CIP~\citep{palermo2022cip}, Emokine~\citep{christensen2024emokine}, UNIPD~\citep{guidolin2022unipd}, and Virginia Natural Motion~\citep{geissinger2020virginianaturalmotion}; 1{,}106 training and 177 held-out test trials). Ground-truth articulation is the Xsens skeleton, mapped to the SMPL joint layout following DynaIP.

\textbf{Synthesizing Gyroscope Readings.} None of these datasets provide raw gyroscope readings. To match the six-axis input format of \Name{}, we synthesize per-sensor angular velocity from the calibrated global-orientation sequence $R_t \in SO(3)$ via the relative rotation between consecutive frames: $\omega_t = \mathrm{Log}(R_t^\top R_{t+1}) / \Delta t$, where $\Delta t = 1/60$\,s and $\mathrm{Log}(\cdot)$ is the $SO(3)$ logarithm map, yielding angular velocity in the sensor frame. The finite difference is computed on the full uncropped sequence so that frames near a later crop retain valid neighbors (the last frame repeats the previous value). Since DynaIP takes the $SO(3)$ sequence directly as input while \Name{} uses the finite difference of the same sequence, the comparison is fair as neither model has access to information unavailable to the other.

\subsection{Model}
The backbone is the pretrained \Name{}-85M encoder, kept \emph{frozen}. Per IMU we encode a 480-frame (8\,s) window into 30 patch tokens (patch size 16). Tokens from the six IMUs are concatenated patch-wise ($6{\times}768$) and projected to 512 dimensions. The MLP probe applies a 2-layer MLP (dim 512), linearly interpolates the resulting sequence back to the input resolution (480 frames), and predicts per-frame 16-joint root-relative 6D rotations of the SMPL skeleton with a per-frame linear head (2.67M trainable parameters). The transformer probe inserts a 2-layer transformer decoder (dim 512, 8 heads, 7.44M trainable parameters) between the projection and the interpolation head: 30 learnable pose queries cross-attend to the projected patch tokens, then follow the same per-patch MLP, interpolation, and per-frame head. Predictions are assembled into 24-joint global rotations using the measured pelvis orientation for the root, measured calibrated orientations for IMU-equipped joints, and parent copying for IMU-less joints, following the assembly of DynaIP.

\subsection{Training}
Decoders are trained with a 6D rotation MSE loss on the Xsens training split (additionally including DIP-IMU for the Xsens+DIP variants), using AdamW (lr 1e-3, weight decay 1e-4, cosine schedule with warmup), batch size 48, 30 epochs, windows of 480 frames with stride 240, and bf16 autocast. Main-text models use gravity-restored accelerometer inputs. For the finetuned variant ($\dagger$ in Table~\ref{tab:exp-mocap}), the backbone is unfrozen and trained jointly with the decoder at a lower learning rate (1e-4).

\subsection{Evaluation Protocol}
The entire test sequence is predicted in one pass with sliding windows stitched at their centers. We use the official DynaIP evaluator and metrics: SIP error (upper-arm/upper-leg global rotation), global angular error (GAE, root and ignored joints replaced by ground truth), joint position error (JPE, cm, via SMPL/Xsens forward kinematics), and mesh vertex error (cm, SMPL only). Results are reported in Table~\ref{tab:exp-mocap} of the main text.

\section{Robot Odometry Experiment Setup}
\label{app:tartan-protocol}

\subsection{Task and Datasets}
Given a 1\,s window of 6-axis IMU data (accelerometer + gyroscope, 200\,Hz, body frame, gravity retained), the model predicts the window-mean body-frame velocity. Training and validation use the TartanIMU challenge corpus (car/dog/drone/human; 395 training and 80 validation trajectories, 27\,h). Labels are window means of the ground-truth body velocity.

\subsection{Model}
The backbone is the pretrained \Name{} encoder kept \emph{frozen}. Each IMU stream is encoded into patch tokens, and the decoder operates on the final-layer features of the input window (12 patch tokens for the default 1\,s input). We consider two decoders: a 2-layer transformer (dim 512, 8 heads, 6.96M parameters) that also consumes the \texttt{CLS} token and mean-pools over tokens, and a lighter MLP probe (3 layers, dim 512, 0.66M parameters) over the mean-pooled patch tokens. Both regress the 3D body velocity and receive no platform information. Training uses Huber loss ($\beta{=}1$), AdamW (lr $10^{-3}$, weight decay 0.05, warmup + cosine), 30 epochs, EMA (0.999), and platform-balanced sampling of training windows.



\subsection{Sliding Window Strategies}
We consider two sliding window strategies: the sparse variant uses non-overlapping 1\,s windows; the dense variant (main-text default) slides the training windows with a 0.2\,s stride (5$\times$ windows). These settings only affect training, and the evaluation protocol (1-second windows, no overlap) is identical to the official TartanIMU Challenge evaluation script. Results are shown in Table~\ref{tab:app-window}.

\begin{table}[htbp]
\centering
\scriptsize
\begin{tabular}{l|cc|cc|cc}
\toprule
\multirow{2}{*}{Window strategy} & \multicolumn{2}{c|}{AVE$\downarrow$} & \multicolumn{2}{c|}{ATE20$\downarrow$} & \multicolumn{2}{c}{Score$\downarrow$} \\
 & MLP & Transformer & MLP & Transformer & MLP & Transformer \\ \midrule
1\,s, non-overlapping & 0.283 & 0.262 & 1.103 & 0.992 & 0.373 & 0.341 \\
1\,s, 0.2\,s stride (default) & 0.249 & 0.209 & 1.012 & 0.854 & 0.333 & 0.280 \\
\bottomrule
\end{tabular}
\caption{Window-strategy ablation on the TartanIMU validation split (frozen \Name{}-85M, macro-averaged over platforms). Denser training windows help both probes.}
\label{tab:app-window}
\end{table}

\subsection{Evaluation Protocol}
Models are scored on the official validation split with the official TartanIMU metrics: per-platform AVE (m/s) and ATE20 (m), macro-averaged over the four platforms, and the official TartanIMU Score ($0.6\cdot\mathrm{AVE}/0.7356 + 0.4\cdot\mathrm{ATE20}/3.116$; lower is better). \Name{} is evaluated platform-agnostic (a single shared model for all platforms). Results are reported in Table~\ref{tab:exp-tartan} of the main text.

\section{Additional MoCap Qualitative Results}
\label{sec:appendix-mocap-visualization}
We present addition visualized results from the motion capture task. \Name{} dense features can support accurate inertial human pose estimation (HPE). With task-specific decoder architectures and training objective with physical and body priors, we expect \Name{} to be a serious contender for SOTA inertial HPE.

\begin{figure*}[htbp]
  \centering
  \includegraphics[width=\textwidth]{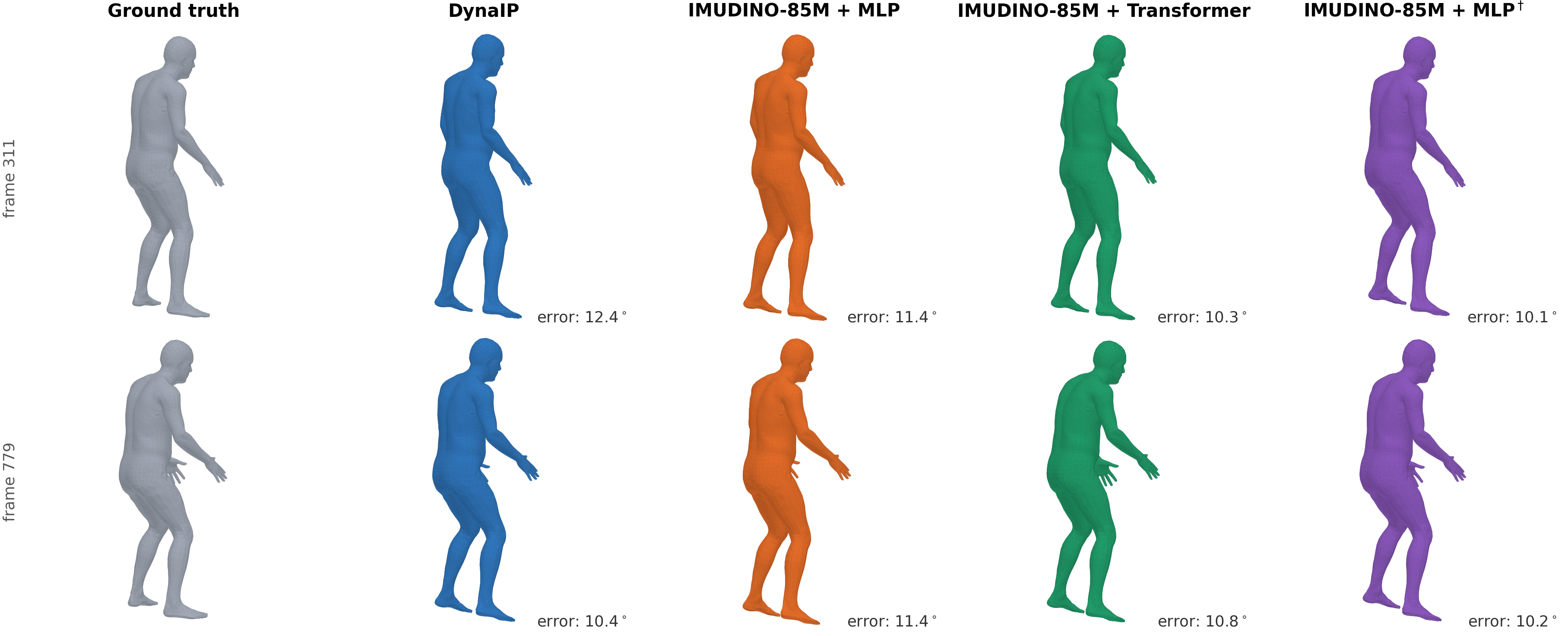}
  \caption{Results on a squatting sequence from the UNIPD test set.}
  \label{fig:app-mocap-squat}
\end{figure*}

\begin{figure*}[htbp]
  \centering
  \includegraphics[width=\textwidth]{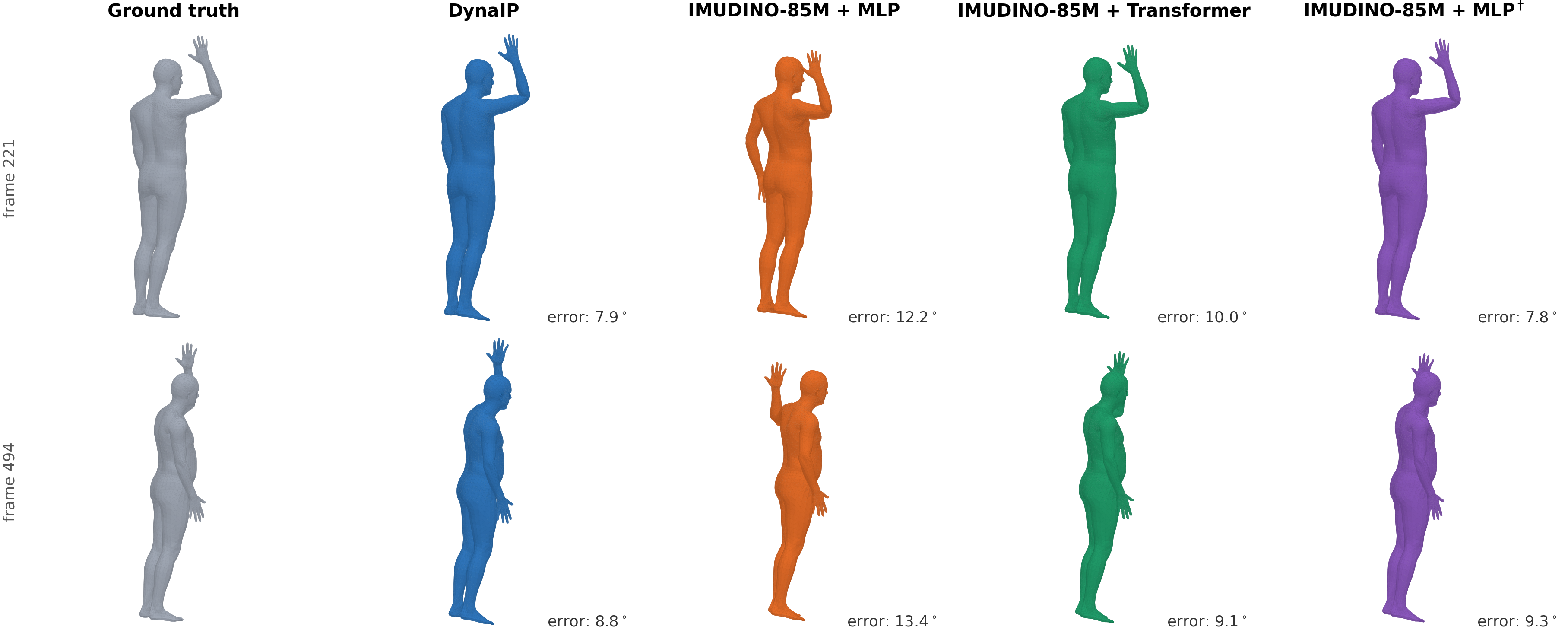}
  \caption{Results on a throwing sequence from the UNIPD test set.}
  \label{fig:app-mocap-throw}
\end{figure*}


\begin{figure*}[htbp]
  \centering
  \includegraphics[width=\textwidth]{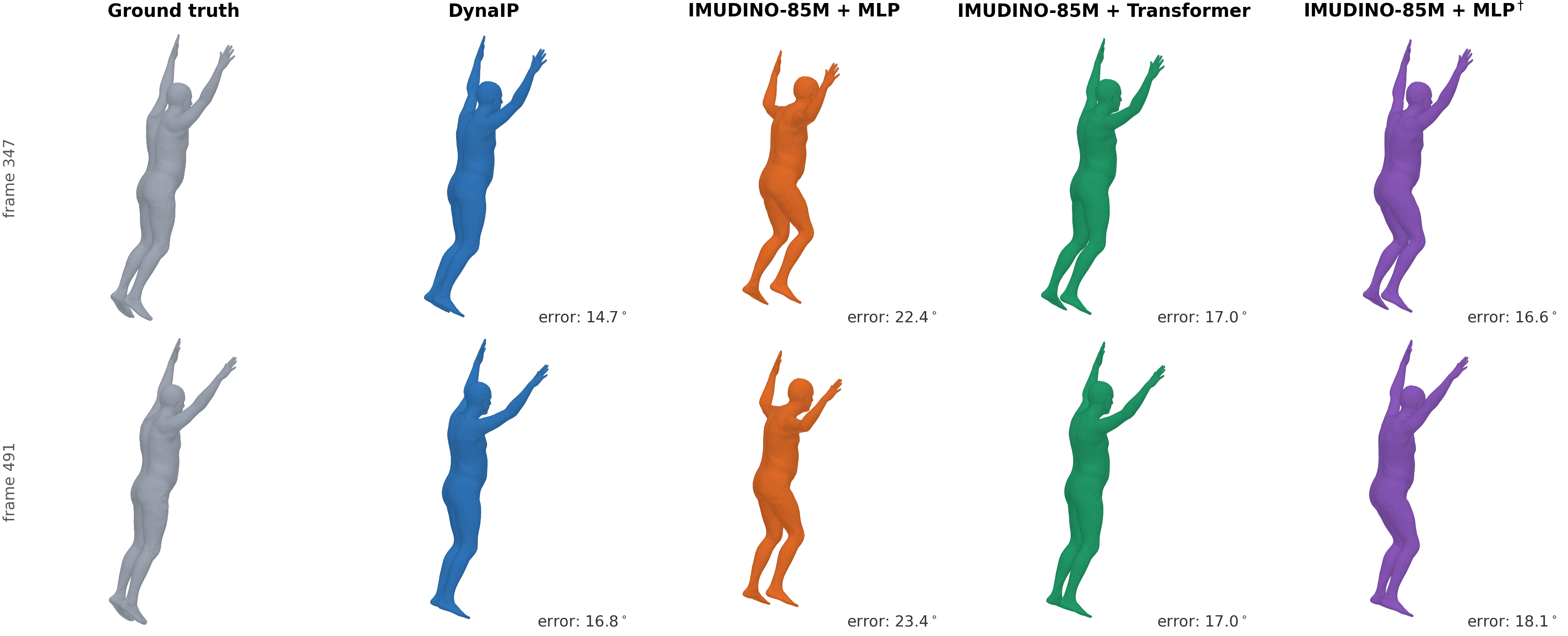}
  \caption{Results on a jumping sequence from the UNIPD test set.}
  \label{fig:app-mocap-jump}
\end{figure*}

\begin{figure*}[htbp]
  \centering
  \includegraphics[width=\textwidth]{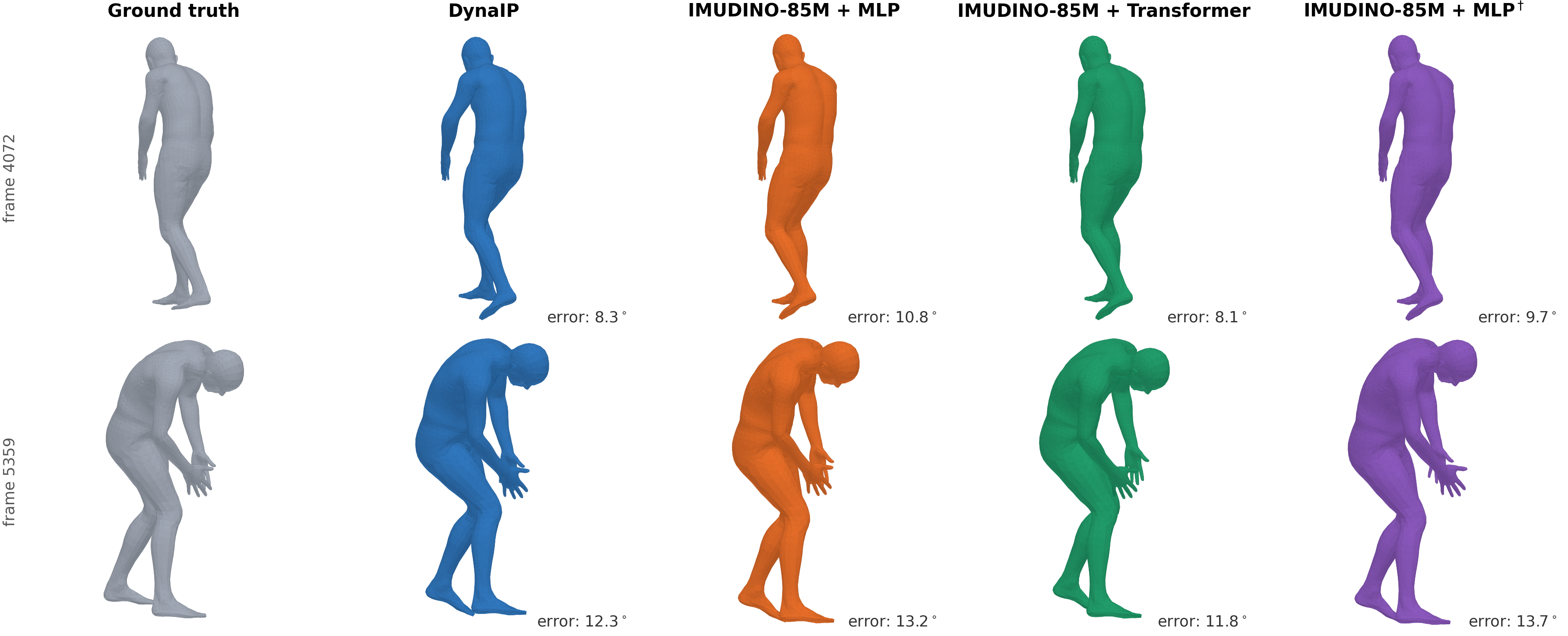}
  \caption{Results on a lifting sequence from the AnDy test set.}
  \label{fig:app-mocap-andy}
\end{figure*}

\begin{figure*}[htbp]
  \centering
  \includegraphics[width=\textwidth]{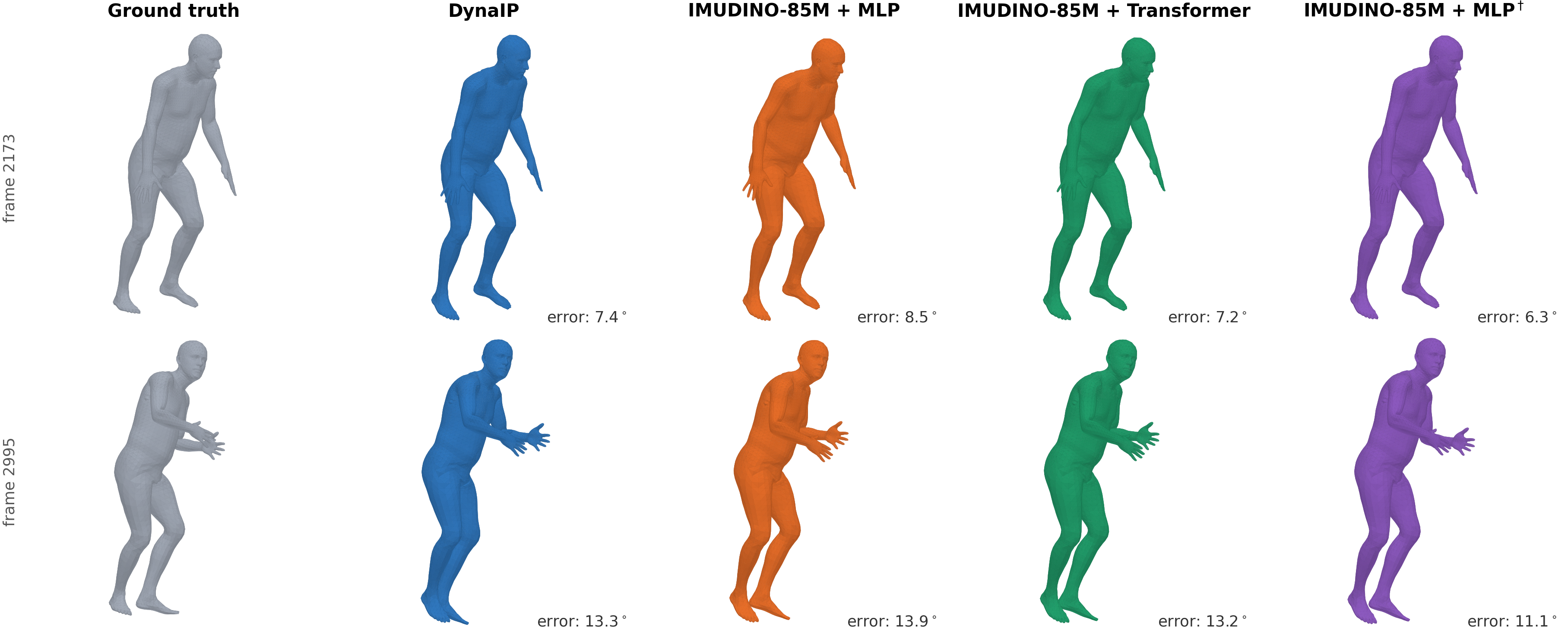}
  \caption{Results on a circuit sequence from the CIP test set.}
  \label{fig:app-mocap-cip}
\end{figure*}

\begin{figure*}[htbp]
  \centering
  \includegraphics[width=\textwidth]{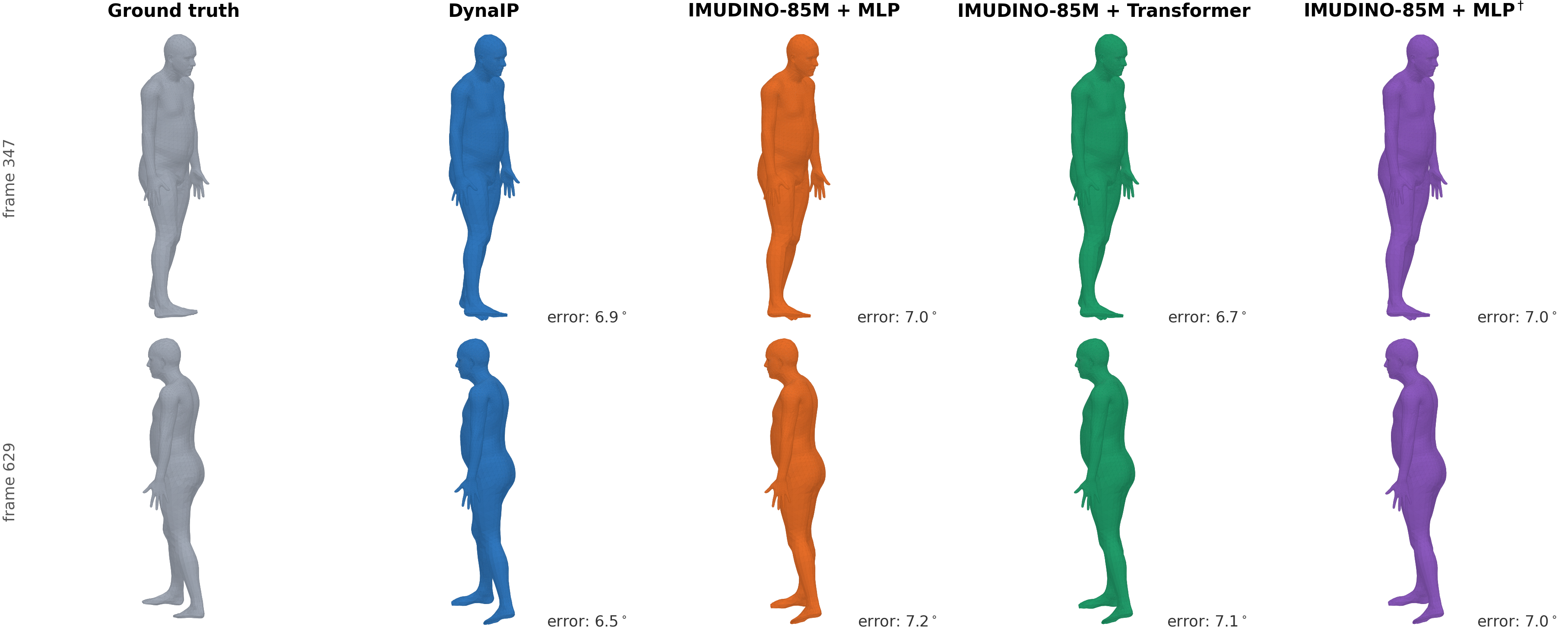}
  \caption{Results on a walking sequence from the UNIPD test set.}
  \label{fig:app-mocap-walk}
\end{figure*}

\end{document}